\documentclass[lettersize,journal]{IEEEtran}
\usepackage{amsmath}
\usepackage{amsfonts}
\usepackage{algorithm}
\usepackage{algorithmicx}
\usepackage{array}
\usepackage[caption=false,font=normalsize,labelfont=sf,textfont=sf]{subfig}
\usepackage{textcomp}
\usepackage{stfloats}
\usepackage{url}
\usepackage{verbatim}
\usepackage{graphicx}
\usepackage{cite}
\usepackage{blindtext}
\usepackage{subfig}

\usepackage{bbm}
\usepackage{amssymb}
\usepackage{subfig}
\usepackage{tabularx}
\usepackage{makecell}
\usepackage{booktabs}
\usepackage{multirow}

\usepackage{picins}
\usepackage{color}

\usepackage{caption}
\begin{document}

\title{Contrast, Imitate, Adapt: Learning Robotic Skills from Raw Human Videos
}

\author{Zhifeng Qian, Mingyu You$^{*}$, Hongjun Zhou, Xuanhui Xu, Hao Fu, Jinzhe Xue and Bin He
\thanks{This work was supported in part by the National Natural Science Foundation of China under Grant No. 62073244 and 61825303, 62088101.}
\thanks{Authors are with the College of Electronic and Information Engineering, Frontiers Science Center for Intelligent Autonomous Systems, Tongji University. M.Y is the corresponding author with {\small myyou@tongji.edu.cn}. }
}



\maketitle

\begin{abstract}
Learning robotic skills from raw human videos remains a non-trivial challenge. Previous works tackled this problem by leveraging behavior cloning or learning reward functions from videos. Despite their remarkable performances, they may introduce several issues, such as the necessity for robot actions, requirements for consistent viewpoints and similar layouts between human and robot videos, as well as low sample efficiency. To this end, our key insight is to learn task priors by contrasting videos and to learn action priors through imitating trajectories from videos, and to utilize the task priors to guide trajectories to adapt to novel scenarios. We propose a three-stage skill learning framework denoted as Contrast-Imitate-Adapt (CIA). An interaction-aware alignment transformer is proposed to learn task priors by temporally aligning video pairs. Then a trajectory generation model is used to learn action priors. To adapt to novel scenarios different from human videos, the Inversion-Interaction method is designed to initialize coarse trajectories and refine them by limited interaction. In addition, CIA introduces an optimization method based on semantic directions of trajectories for interaction security and sample efficiency. The alignment distances computed by IAAformer are used as the rewards. We evaluate CIA in six real-world everyday tasks, and empirically demonstrate that CIA significantly outperforms previous state-of-the-art works in terms of task success rate and generalization to diverse novel scenarios layouts and object instances.

\end{abstract}

\def\abstractname{Note to Practitioners}
\begin{abstract}
This work aims to study robot skill learning from raw human videos. Compared with teleoperation or kinesthetic teaching in the laboratory, such learning method can flexibly utilize large-scale human videos available on the Internet, thereby improving the robot's ability to generalize to various complex scenarios. 
Previous works on learning from videos usually have some issues, including requirements for robot actions, consistent viewpoints, similar layouts and low sample efficiency.
To alleviate these issues, we propose a three-stage skill learning framework CIA.
Temporal alignment is utilized to learn task priors through our proposed transformer-based model and self-supervised loss functions.
A trajectory generation model is trained to learn the action priors.
To further adapt to diverse scenarios, we propose a two-stage policy improvement method by initialization and interaction. An optimization method is introduced to ensure safe interaction and sample efficiency, where the optimization objective is guided by the learned task priors.
The experimental results show that our CIA outperforms other state-of-the-art methods in task success rate and generalization to novel scenarios.
\end{abstract}


\section{Introduction}
\IEEEPARstart{I}{ntelligent} agents are expected to master various general-purpose manipulation skills in an open environment. Towards this goal, Reinforcement Learning (RL) and Imitation Learning (IL) have recently made advanced progress in many tasks such as grasping \cite{10155203}, pouring \cite{zhang2022one}, and completing assembly tasks \cite{li2023using}. 
Despite these remarkable advancements, they cannot scale to various real-world scenarios outside the laboratory.
Specifically, RL methods \cite{10155203} are usually restricted in simple simulations, which leads to insufficient generalization ability to diverse real-world scenarios.
IL methods \cite{zhang2021explainable} perform supervised learning on demonstration trajectories collected by teleoperation or kinesthetic teaching. However, such data collection is time-consuming and labor-intensive, which is limited to laboratory settings. 
In contrast, human videos can be easily captured by different people in various scenarios, and fortunately, there are already a large number of videos on the Internet, which contain rich viewpoints, operating embodiments, diverse objects, etc. 
Therefore, we delve into the question of how to learn robotic skills from underutilized raw human videos (e.g. YouTube videos) and improve generalization.

\begin{figure}[t]
	\centering
	\includegraphics[width=8cm]{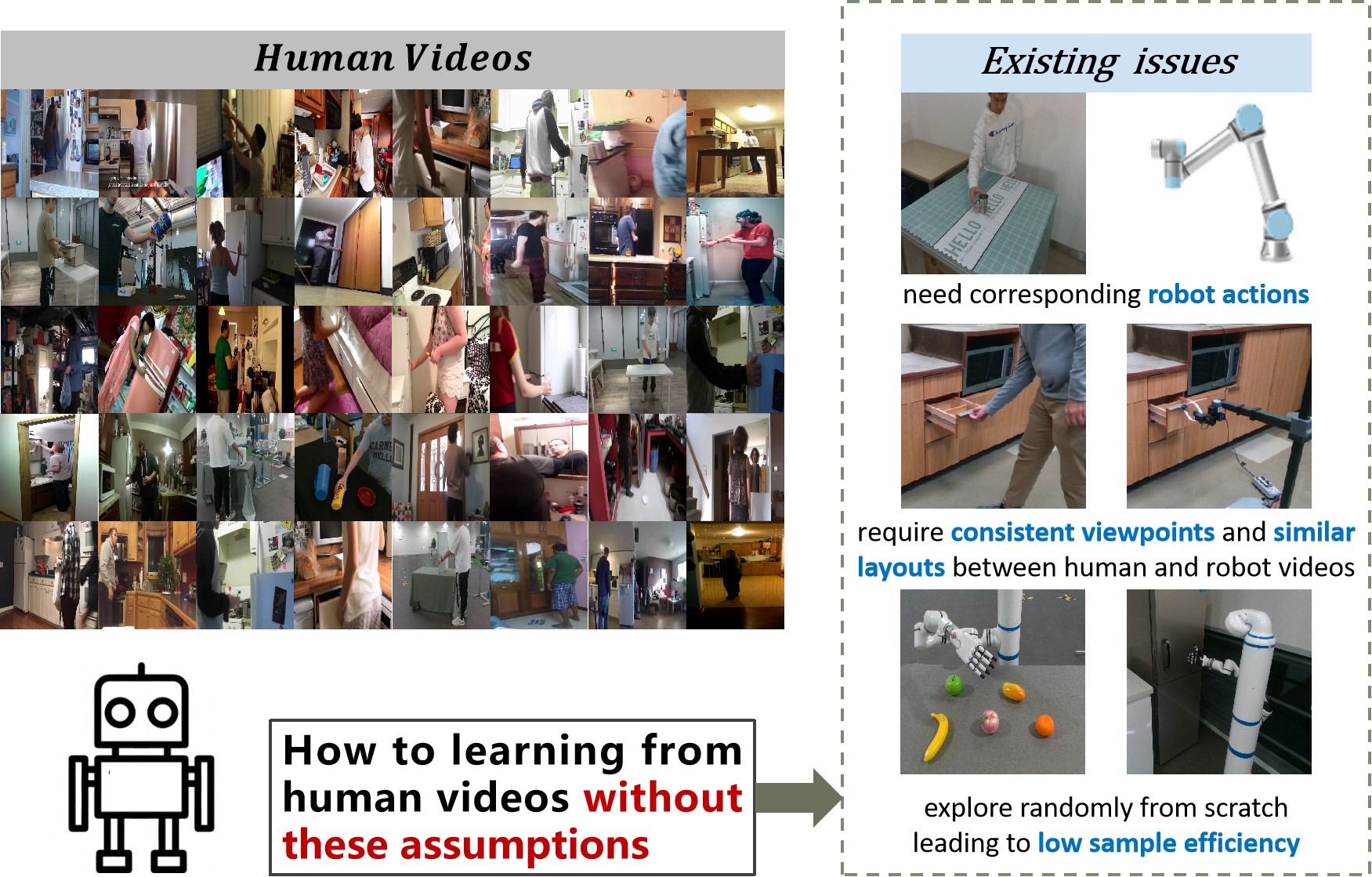}
	\caption{Existing issues of robot learning from human videos. 
	}
	\label{intro}
	\vspace{-1.6em}
\end{figure}

Many works attempt to learn from videos, but have some issues, including introducing unreasonable assumptions about videos or low sample efficiency, which are described in Fig. \ref{intro}. 
To be specific, some IL methods utilize Behavior Cloning algorithms \cite{zitkovich2023rt} to learn pixel-to-action mapping.
For instance, Zhang et al. \cite{zhang2022one} efficiently learn to pour drinks from videos while our previous work \cite{qian2023robot} proposes viewpoint transformation to deal with inconsistent context issues of human demonstrations.
However, these methods require the corresponding action labels, which is impractical for raw human videos and may limit the scalability of IL beyond laboratory settings.

Other RL methods learn a reward function, e.g. a video classifier \cite{shao2021concept2robot}, an embodiment-agnostic representation \cite{bahl2022human,Zhou-RSS-21} or disentangled representations of objects \cite{qian2023goal,qian2022weakly}, which can measure the similarity between robot execution videos and human videos. The rewards are used to make robots perform visually the same actions as humans.
However, they require consistent viewpoints and similar layouts between human videos and robot execution videos, which imposes great restrictions on human videos and testing scenarios. 
In addition, instead of leveraging action priors in human videos to speed up exploration, most RL methods randomly explore from scratch in a vast action space, which leads to low sample efficiency and limits the flexibility of their methods.

The pursuit of efficiently learning robotic skills from human videos and generalizing them to novel scenarios without the above issues remains a non-trivial challenge.
To this end, we propose a three-stage skill learning framework denoted as \textbf{Contrast-Imitate-Adapt (CIA)}. 
Our key insight is to learn task priors by \textbf{contrasting} videos and action priors through \textbf{imitating} trajectories from videos, and then use task priors to guide actions to \textbf{adapt} to novel scenario layouts and object instances.
To be specific, we propose an interaction-aware alignment transformer (\textbf{IAAformer}) to temporally align video pairs to understand the hand-object interaction process.
Attention mechanisms and a hindsight interaction contrastive loss are designed to improve the alignment effect. 
Then CIA distills the action priors from videos into a trajectory generation model as our policy.
To adapt to novel scenarios, we utilize RL to improve the policy. However, it is dangerous and inefficient to interact with the real environment. Therefore, we propose an \textbf{Inversion-Interaction} method to initialize coarse trajectories by GAN inversion and refine them with limited interactions. In particular, we introduce an optimization method based on semantic directions sampling of trajectories for safe interaction and high sampling efficiency. Moreover, the alignment distance computed by IAAformer can serve as the reward function for the proposed optimization method.

We perform extensive experiments on six different real-world manipulation tasks on a robotic arm-hand system. We analyze CIA in terms of success rate, effectiveness, and generalization compared to state-of-the-art baselines.
The main contributions of this paper are as follows:
\begin{itemize}
	\item We introduce a three-stage imitation learning framework CIA that enables agents to learn skills from raw human videos and adapt to novel scenarios.
	\item We propose IAAformer with self- and cross-attention mechanisms and a hindsight interaction contrastive loss, which can temporally align two videos and compute alignment distance as a subsequent reward function.
	\item We design a two-stage Inversion-Interaction Method to obtain appropriate trajectories from coarse initialization to refined interaction improvement. Based on trajectory semantic directions in the latent space, we introduce an optimization method to safely and efficiently improve the policy.
	\item We empirically demonstrate that our CIA can learn robotic skills from Internet human videos and outperform other state-of-the-art methods in six real-world tasks.

\end{itemize}

The organization of this paper is as follows:
Related works are listed in Section \ref{section2}. 
Our proposed framework CIA is introduced in Section \ref{section3}. 
In Section \ref{section4}, extensive experimental details and analysis are provided.
The conclusion is summarized in Section \ref{section5}.

\section{RELATED WORK}\label{section2}

\subsection{Imitation and Reinforcement Learning from Videos}

\noindent\textbf{Imitation Learning from Videos }
Imitation learning (IL) is a promising method to learn robotic skills from demonstrations.
To learn from videos, many works propose Behavior Cloning methods to learn the mapping from visual inputs to robot actions.
Zitkovich et al. \cite{zitkovich2023rt} finetune a pre-trained Vision-Language Model on large-scale robot data, enabling strong generalization for many robot tasks.
Our previous work \cite{qian2023robot} learns skills from human demonstrations with inconsistent contexts 
while Wang et al. \cite{wang2023mimicplay} learn latent plans from videos to perform long-horizon imitation learning.
Zhang et al. \cite{zhang2022one} propose an one-shot domain-adaptive imitation learning framework to adapt to new scenarios. 
However, these methods assume access to the corresponding robot actions, which is challenging to achieve in practical settings.

\noindent\textbf{Learning Rewards from Videos }
To alleviate the issue of lacking robot actions, many works learn various reward functions from offline datasets (e.g. videos) to provide Reinforcement Learning (RL) algorithms. 
Our previous work \cite{qian2022weakly} learns a disentangled representation in which different parts correspond to different object attributes (position, shape, color) while some works \cite{bahl2022human, Zhou-RSS-21} learn an embodiment-agnostic representation.
These learned representations can be used to compute the rewards of RL to reproduce the behavior in the videos.
Many other works \cite{shao2021concept2robot, nair2022learning} 
train video classifiers to determine if the task performed by robots is the same as in the expert videos as the reward.
However, these methods necessitate consistent viewpoints or similar layouts between human videos and robot execution videos, potentially imposing constraints on the flexibility of IL.
In contrast, we aim to learn skills from human videos without these assumptions. 

\noindent\textbf{Integrating Reinforcement Learning with Imitation Learning
} Many works propose to integrate IL with RL to improve the learning efficiency. Dey et al. \cite{dey2021joint} propose a control criterion to enable the RL policy from imitating the baseline policy to gradually learning an optimal policy. However, it is inflexible to assume that there is a baseline policy. Hafez et al. \cite{hafez2023continual, hafez2021behavior} learn behavior embeddings as task representations from self-organization demonstrations, which are used to perform more efficient RL. Kalashnikov et al. \cite{kalashnikov2021mt} propose a robotic learning system to efficiently learn a wide range of skills by sharing experiences and representations across tasks. Eysenbach et al. \cite{eysenbach2022contrastive} use contrastive learning to make the representations correspond to a value function, leading to a much simpler but efficient goal-conditioned RL method. Although these methods use task or state representations to improve the efficiency, a large part of the inefficiencies are caused by huge action spaces, as mentioned in \cite{shaw2023videodex}. In contrast, we extract human trajectories for action prior learning, which improves the efficiency of RL learning more directly and efficiently.

\noindent\textbf{Learning from Large-Scale Human Datasets }
There are many large-scale datasets of human activities with crowd-sourced annotation, e.g. 
100 Days of Hands \cite{shan2020understanding} for hand-object interactions, Dex-YCB \cite{chao2021dexycb} for grasping of diverse objects. VLOG \cite{fouhey2018lifestyle} is collected from YouTube videos uploaded by various people documenting their daily lives. Charades \cite{sigurdsson2016hollywood} is collected with hundreds of people recording everyday activities in their homes. 
In our work, we utilize videos in Dex-YCB \cite{chao2021dexycb}, VLOG \cite{fouhey2018lifestyle} and Charades \cite{sigurdsson2016hollywood} to train our framework.

\begin{figure}[t]
	\centering
	\includegraphics[width=8.8cm]{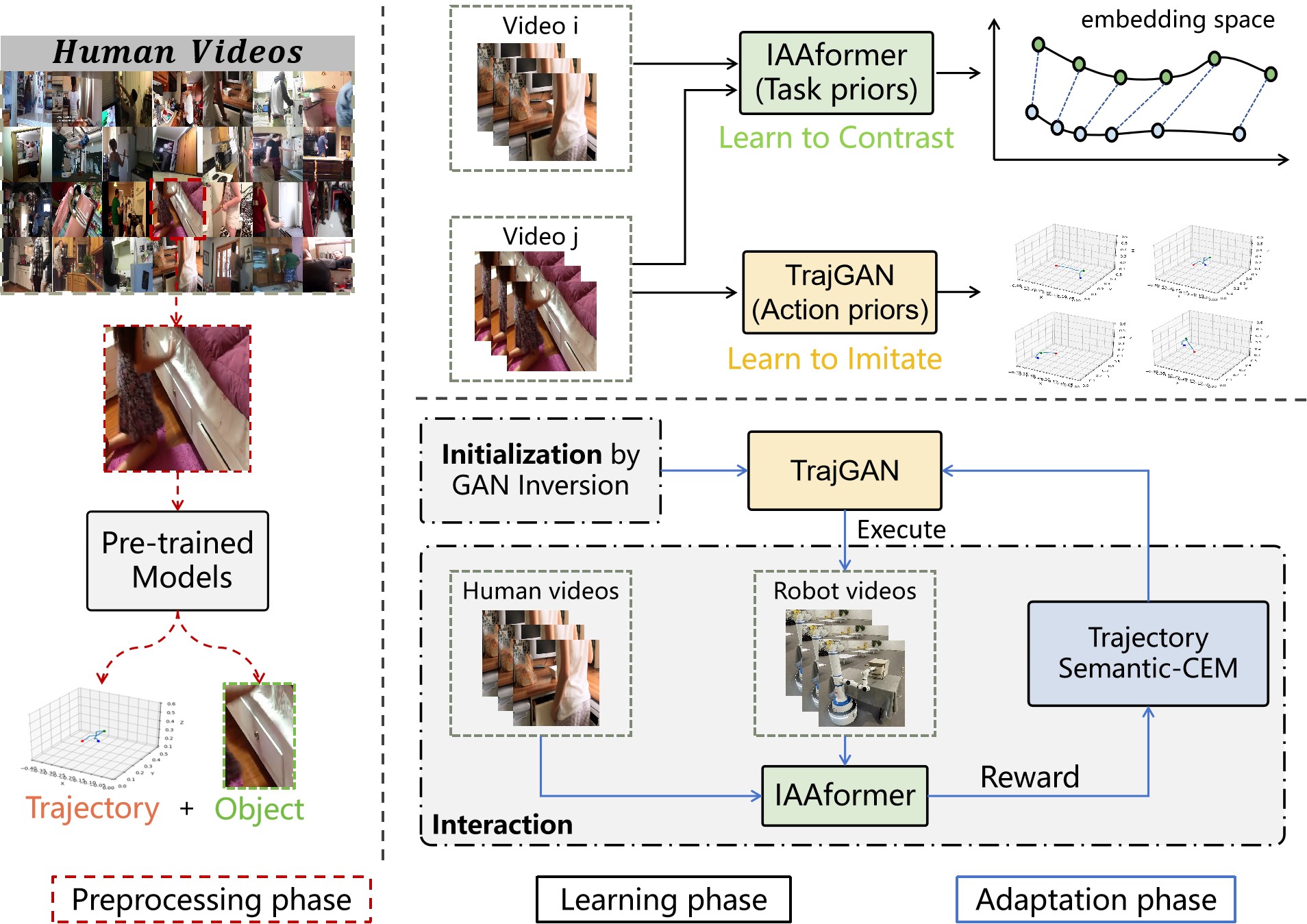}
	\caption{Illustration of our framework CIA. Priors are first extracted by pre-trained models. Then, IAAformer aims to temporally align video pairs while TrajGAN learns to generate trajectories. 
	To adapt to novel scenarios, TrajGAN is initialized by GAN inversion and improved by the proposed TS-CEM where the rewards are output by IAAformer.
	}
	\label{method-fig}
	\vspace{-1.6em}
\end{figure}

\subsection{Temporal Alignment in Self-supervised Representation Learning}

Self-supervised temporal alignment has been recently proposed to find per-frame correspondences between two videos, which can effectively capture human activity phases and be used for downstream action learning.
Sermanet et al. \cite{Sermanet2017Time} employ a time-contrastive network (TCN) on time-synchronized videos, despite different viewpoints. 
Dwibedi et al. \cite{dwibedi2019temporal} propose a temporal cycle consistency (TCC) loss to maximize the embedding distance between matched pairs of two videos. 
Sanjay et al. \cite{haresh2021learning} and Weizhe et al. \cite{liu2022learning} adopt Dynamic Time Warping (DTW) as a differential alignment loss. 
However, the above methods only focus on learning image representations.
Relying on the advances in computer vision, Kwon et al. \cite{kwon2022context} extract 3D human keypoints from the human activity images and learn the temporal alignment on these skeletons. 
However, this method can only learn skeletal representations, which ignore the information of objects interacting with humans and cannot understand more complex hand-object interaction processes.

\subsection{Sample-Efficient Reinforcement Learning}

The issue of low sampling efficiency is a key limitation of Reinforcement Learning (RL) algorithms. To mitigate this issue, our previous work \cite{qian2023goal} proposes a reachability discrimination module, which is utilized to search for the most reachable sequence of subgoals to improve the sample efficiency of RL for solving temporally extended tasks. 
Curiosity-motivated exploration \cite{Savinov2019_EC} for RL is utilized to encourage agents to visit various states.
For quick optimization and convergence, gradient-free Cross-Entropy Method (CEM) and variants \cite{de2005tutorial, bahl2022human} are utilized to improve the sample efficiency. 
In our work, we modify CEM with our trajectory semantic directions in Section \ref{sec-3.4.3} to interact securely with the real world and converge quickly.

\section{Learning Skills by Contrast-Imitate-Adapt}\label{section3}
We propose a novel three-stage robot learning framework comprising Contrast, Imitation and Adaptation (CIA), which can learn robotic skills from raw human videos and generalize to diverse scenarios. The overall framework CIA is shown in Fig. \ref{method-fig}. 
First, we utilize recent advancements in computer vision to predict 3D human keypoints and object bounding boxes as human priors and convert to the learnable priors, which is introduced in Section \ref{section-3.1}. 
Then the proposed interaction-aware alignment transformer (IAAformer)  learns temporal alignment as task priors through attention mechanisms and a hindsight interaction contrastive loss, which is presented in Section \ref{section-3.2}. 
And TrajGan is used as the policy to learn action priors. 
Finally, to adapt to diverse novel scenarios, we propose an Inversion-Interaction method to initialize coarse trajectories by GAN inversion and refine them with limited interactions, which is shown in Section \ref{section-3.4}.
\subsection{Priors Extraction and Transformation}\label{section-3.1}

\begin{figure}[t]
	
	\label{fig-extract}
	\centering
	\subfloat[Extraction and transformation of human priors.]{\small
		\begin{minipage}{8.cm} 
			\label{fig-extract-a}
			\includegraphics[width=\textwidth]{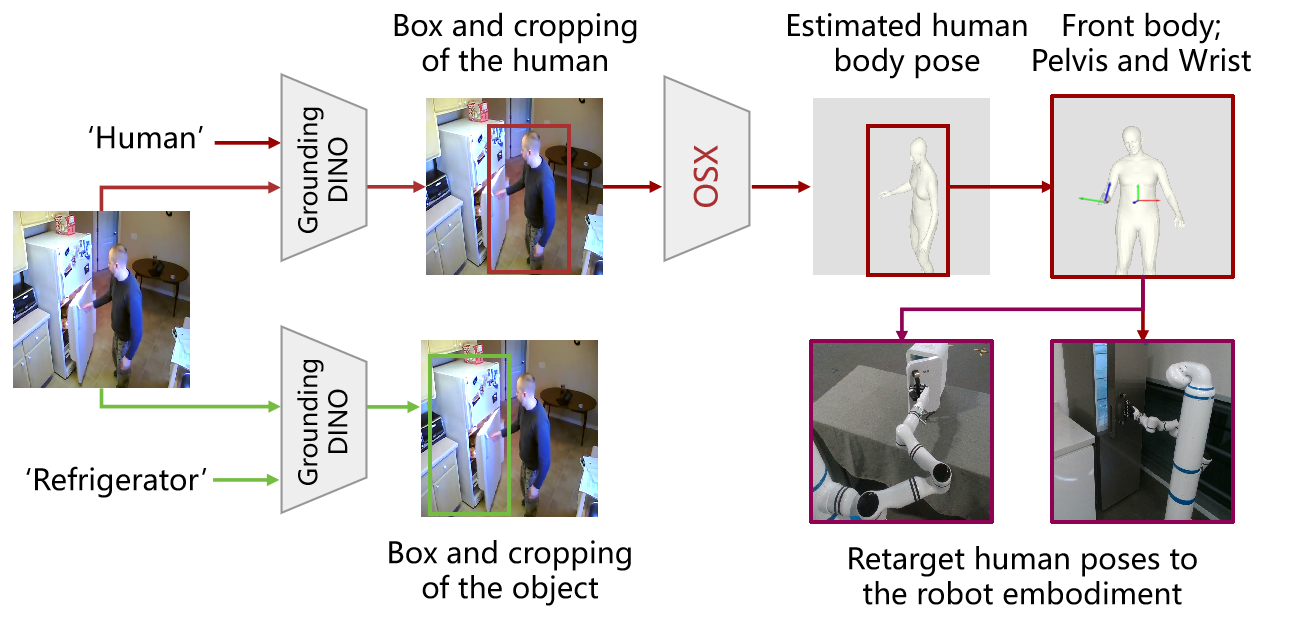} \\
			\vspace{-1.em}
		\end{minipage}
	}
	
	\subfloat[Examples of action priors.]{
		\begin{minipage}{8.2cm}
			\label{fig-extract-b}
			\includegraphics[width=\textwidth]{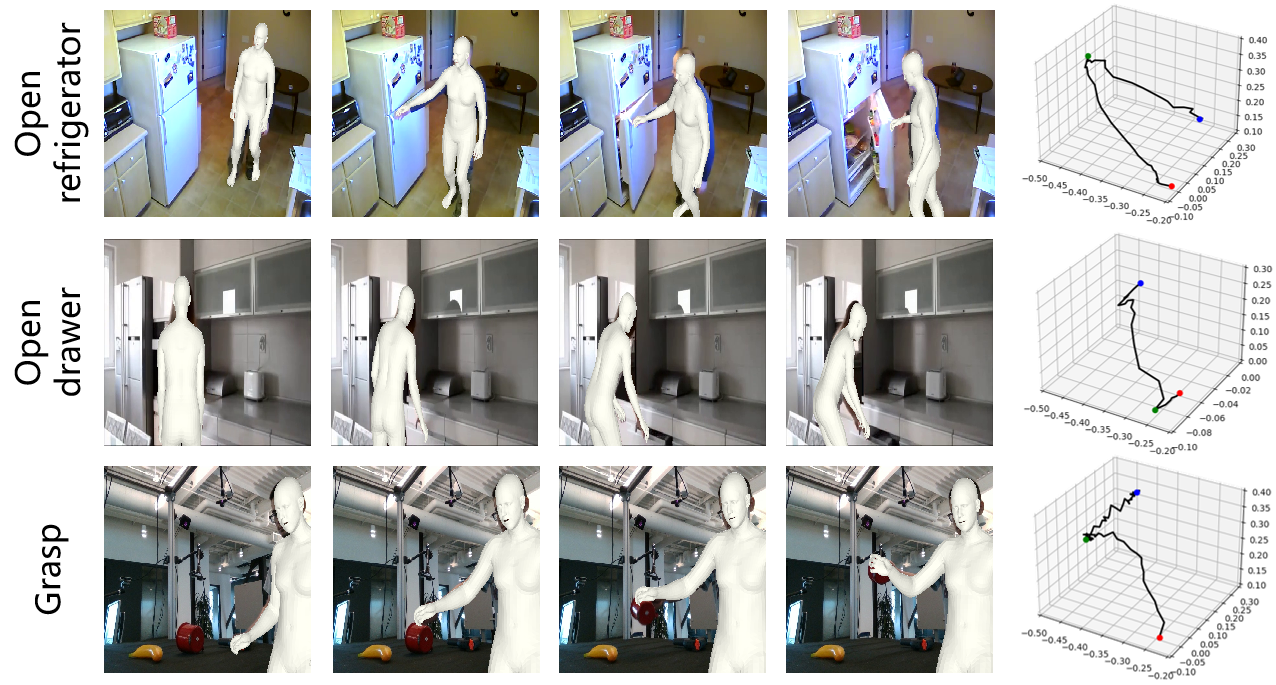} \\
			\vspace{-1.em}
		\end{minipage}
	}
	\caption{Illustration of extraction and transformation of human priors and examples of action priors. } 
\end{figure}

\noindent\textbf{Extracting Human Priors}
We leverage off-the-shelf advances in computer vision to extract human priors from videos. The process is shown in Fig. \ref{fig-extract-a}.
Given each frame $V^h_i$ of the human video $V^h$, we use an open-set object detector Grounding DINO \cite{liu2023grounding} to detect the bounding box of the task-relevant objects $B_{obj}$. Then, a 3D human mesh recovery model OSX \cite{lin2023one} is utilized to estimate the joint rotation $\theta_{body}$ of 3D human model SMPL-X \cite{pavlakos2019expressive}, which can further compute 3D keypoints $P_{3D}$ and 2D keypoints $P_{2D}$. We use the first wrist rotation to describe the hand rotation $\theta_{wrist}$.
In addition, we employ a contact detection model \cite{shan2020understanding} built on top of Faster-RCNN \cite{ren2015faster} to predict the binary hand-object contact state $c$. Since single frame detection may be noisy, we use Savitzky-Golay filtering \cite{savitzky1964smoothing} to smooth across timesteps.

\noindent\textbf{Transforming to Learnable Action Priors }
Although we extract human priors, we lack a fixed anchor to retarget trajectories in videos from different viewpoints into the same coordinate system, e.g. the robot coordinate system.
Therefore, such trajectories cannot be directly learned by robots.
Previous methods imposed strict consistency constraints on viewpoints \cite{bahl2022human} or involved complex estimation of wrist-to-world coordinate mappings \cite{shaw2023videodex}, which leads to cumulative errors.

To this end, we 
use the Pelvis joint of the SMPL-X model \cite{pavlakos2019expressive} as the human anchor point, and select the fixed vertical offset point of the robotic arm base as the robot origin. 
We believe that the relative movement between the human wrist and its own anchor point is similar to the relative movement between the end effector of a robotic arm and the robot origin. 
The specific process is shown in Fig. \ref{fig-extract-a}.
First, to mitigate the influence of diverse viewpoints, we set the root body rotation to zero. 
Then, we obtain the relative position $P_{3D}^{'}$ and rotation $\theta_{wrist}^{'}$ from the Right Wrist joint to the Pelvis joint through the forward calculation of the human kinematic chain. The relative trajectories, named action priors, can be directly deployed to robots and utilized to train TrajGAN in Section \ref{section-3.4}. 
Some examples of action priors are shown in Fig. \ref{fig-extract-b}.
Overall, we obtain priors containing 3D keypoints $P_{3D}^{'}$, the wrist rotation $\theta_{wrist}^{'}$, 2D keypoints $P_{2D}$, the contact state $c$, the object bounding box $B_{obj}$ and corresponding cropped images $I_{obj}$. 

\subsection{Learning Task Priors through Temporal Alignment}\label{section-3.2}

The goal of IAAformer is to learn task priors by performing temporal alignment to identify frame-by-frame consistency between two videos of the same task.
Then, IAAformer can contrast robot execution sequences and human videos to measure the robot's execution quality. 
IAAformer explicitly models the hand-object interaction through self- and cross-attention mechanisms. 
Through our hindsight interaction contrastive loss, we can improve the representation awareness of the interaction process. 
In addition, a proposed temporal monotonic loss ensures the alignment monotonicity.

\vspace{5pt}
\subsubsection{Architecture of IAAformer}\label{sec-3.2.1}
\

\begin{figure}[t]
	\centering
	\includegraphics[width=8.8cm]{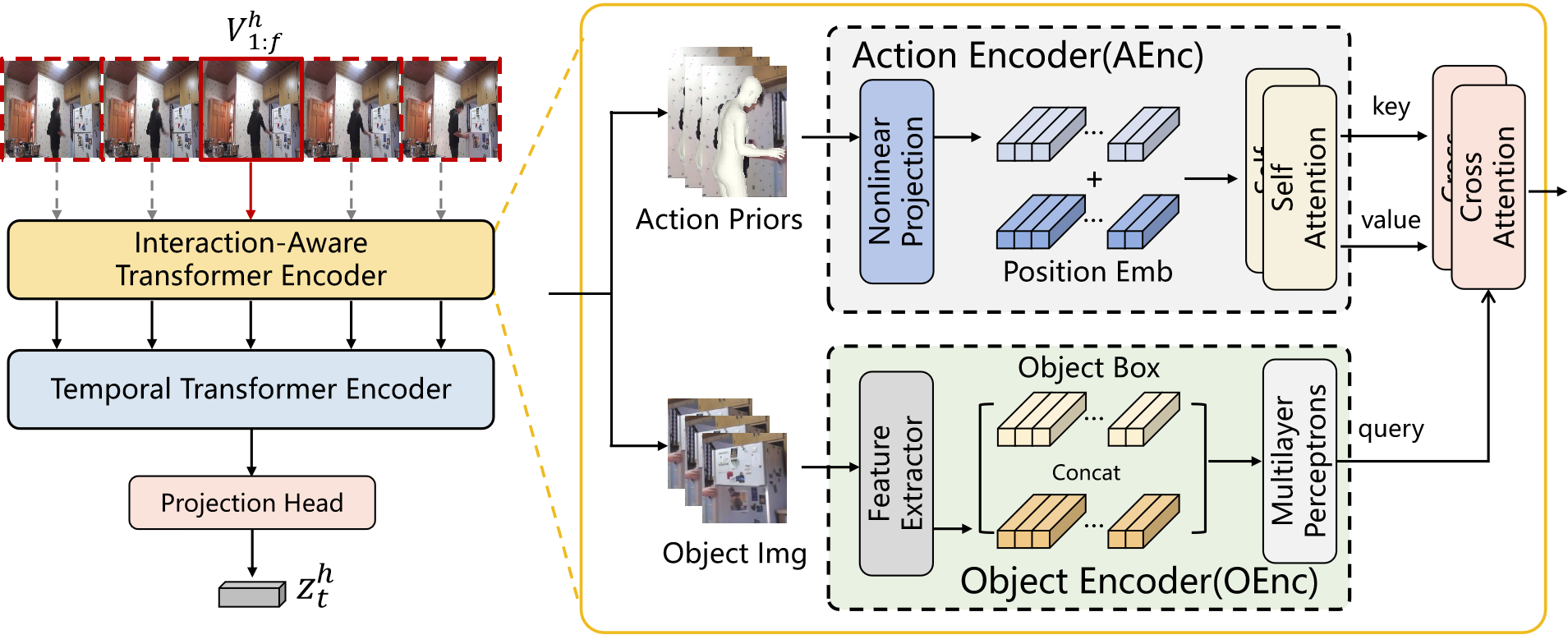}
	\caption{Architecture of our Interaction-Aware Alignment transformer (IAAformer). 
}
	\label{fig-3-2}
	\vspace{-1.em}
\end{figure}

IAAformer is composed of three modules, which are shown in Fig. \ref{fig-3-2}. Interaction-Aware Transformer Encoder is used to model the hand-object interaction with self- and cross-attention mechanisms while Temporal Transformer Encoder focuses on modeling global dependencies across the frames, yielding effective spatiotemporal and contextual cues within the input sequence. A Projection Head is used to merge frame-level features and output the 128-dimensional embedding $z_t^h$.

Given a frame $V_t$ and its adjacent frames in the receptive field, the action priors of each frame (including 2D keypoints $P_{2D}$, 3D keypoints $P_{3D}^{'}$ of the human wrist, rotation $\theta_{wrist}^{'}$ and the contact state $c$) are extracted as well as object information ($B_{obj}, I_{obj}$) as described in Section \ref{section-3.1}. 
Unlike previous works that only model image features \cite{sermanet2018time,dwibedi2019temporal} or 3D human skeletons \cite{kwon2022context}, IAAformer takes both action priors and object information as inputs.

\vspace{3pt}
\noindent\textbf{Action Encoder (AEnc)} is used to extract internal features of the action.
A nonlinear projection layer encodes the 2D and 3D action priors extracted in the preprocessing phase to $C$-dimensional features.
Then learnable positional embedding $z_{pos} \in \mathbb{R}^{f \times C}$ is utilized to retrain the position information across the sequence. 
After adding the two, $z_0 \in \mathbb{R}^{f \times C}$ is obtained and fed to the self-attention module like Transformer Encoder \cite{dosovitskiy2020image, vaswani2017attention}, which models the relations between different action priors. 
Detailed, $z_0$ is first transformed linearly and split into the query matrix $Q$, the key matrix $K$ and the value matrix $V$.
Then the self-attention mechanism is computed by the dot-product operation as denoted in Equ. \ref{equ-3.3}. Finally, the features are encoded by multilayer perceptron (MLP) as $z_{action} \in \mathbb{R}^{f \times C}$.
\begin{equation}
\label{equ-3.3}
Self\_Attention(Q,K,V) = Softmax(QK^T/\sqrt{d})V
\end{equation}

\vspace{1pt}
{\noindent\textbf{Object Encoder (OEnc)} is proposed to capture changes in objects over time. OEnc takes as input the cropped object images $I_{obj}$ and extracts features by the pre-trained Resnet50 \cite{he2016deep}. 
Then the features are concatenated with the features transformed from the object bounding box $B_{obj}$. OEnc finally outputs the object features $z_{obj} \in \mathbb{R}^{f \times C}$ through MLP.}
\begin{equation}
\begin{aligned}
\label{equ-3.5}
Cross\_At&tention(Q_{obj},K_{action},V_{action}) \\
&= Softmax(Q_{obj}K_{action}^T/\sqrt{d})V_{action}
\end{aligned}
\end{equation}

\vspace{3pt}
\noindent\textbf{Cross-Attention mechanisms} are designed to explicitly model the interaction process between hands and objects. The object features $z_{obj}$ serve as object queries $Q_{obj} \in \mathbb{R}^{f \times C}$ to calculate multi-head attention with action keys and values.
The multi-head attention is computed by Equ. \ref{equ-3.5} and outputs $z_{1}\in \mathbb{R}^{f \times C}$. With such an interaction mechanism, we introduce the object changes into the action representations, which is crucial for subsequent temporal alignment.

\noindent\textbf{Temporal Transformer Encoder} is designed to model global temporal dependencies of the input sequence rather than the hand-object spatial interactions. After employing the 1D convolutional layer on the above features $z_{1}$, we apply the standard multi-head self-attention blocks and MLP blocks following \cite{dosovitskiy2020image}. The output of Temporal Transformer Encoder is $Y \in \mathbb{R}^{f \times C}$. Since IAAformer is expected to extract features of an intermediate frame based on the input sequence, a projection head with a layer normalization is applied to merge multiple frame features and output 128-dimensional $z_t^h$.

\vspace{5pt}
\subsubsection{Matching Loss}
\

\begin{figure}[t]
	\centering
	\includegraphics[width=8.8cm]{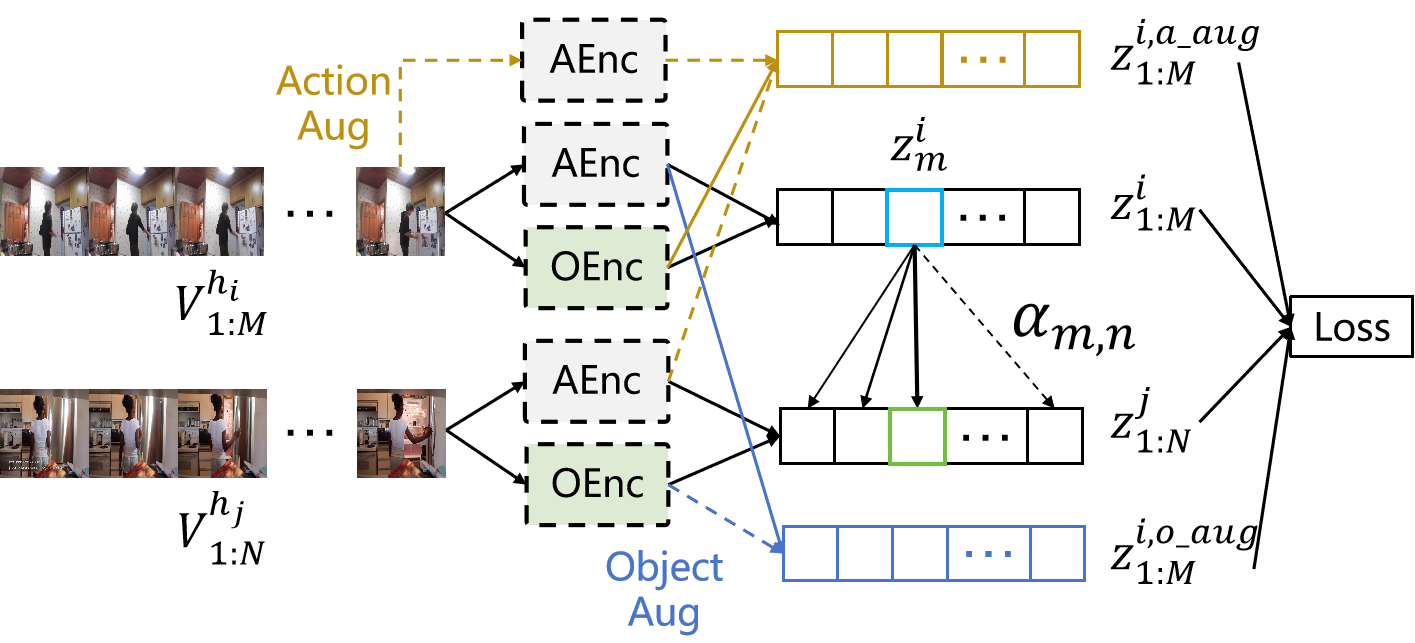}
	\caption{The schematic diagram of self-supervised learning with the hindsight augmentation. }
	\label{fig-aug}
	\vspace{-1.5em}
\end{figure}

We use self-supervised losses to learn the interaction-aware representations. The training process is shown in Fig. \ref{fig-aug}.
Given two videos $V^{h_i}, V^{h_j}$ of the same task, where the length is $M$ and $N$, respectively. Each frame is sent to AEnc and OEnc and encoded respectively as $z_{action}$ and $z_{obj}$, which forms $z_m$ by the subsequent modules of IAAformer. 

Since the frame-level temporal correspondences between two videos are unknown, we utilize temporal cycle-consistency (TCC) loss \cite{dwibedi2019temporal} to mitigate this issue. First, for any frame feature $z_m^i$ in $V^{h_i}$, we find the nearest neighbor in $V^{h_j}$, denoted as $\hat{z_n^j}=argmax_{n \in  [1:N]} sim(z_m^i , z_n^j)$. Then, we determine the nearest neighbor of $\hat{z_n^j}$ in $V^{h_i}$, i.e. $z_k^i$. The matching is cycle-consistent only if $k=m$. 
Therefore, the matching problem is transformed into the classification problem. 
\begin{equation}
\label{equ-3.6}
\hat{z_n^j}=\sum_{k=1}^{N}\alpha_{m,k}z_k^j 
\end{equation}
\begin{equation}
\label{equ-3.7}
\alpha_{m,n} = \frac{exp(  sim(z_m^i ,  z_n^j) / \lambda_{temp} )}{ {\textstyle \sum_{k=1}^{N}exp(sim(z_m^i ,  z_k^j) / \lambda_{temp})} } 
\end{equation}

To enable gradient backpropagation, the softmax function is used to calculate the soft nearest neighbor for $z_m^i$, denoted as Equ. \ref{equ-3.6}. $\alpha_{m,n}$ in Equ. \ref{equ-3.7} is the probability that the $m$-th frame of video $V^{h_i}$ matches the $n$-th frame of video $V^{h_j}$. Let $sim(u, v)=u^Tv/\| u\| \|v\|$ denote the cosine similarity between $u$ and $v$.
We add a temperature parameter $\lambda_{temp}$ to increase the similarity of similar features. 
The classification logits are $\hat{y_m}=softmax(sim(\hat{z_n^j} , z^i_{m = [1:M]}))$. And the ground truth $y_m$ is a one-hot vector while the $m$-th dimension is set to 1.
Finally, the TCC loss is optimized by Equ. \ref{equ-loss1}.
\begin{equation}
\label{equ-loss1}
L_{tcc} = -\sum_{m=0}^{M}y_m log(\hat{y_m})
\end{equation}

However, when trained only on successful interaction videos, TCC loss cannot enable IAAformer to effectively capture the task-relevant structural information in the images.
During task execution, minor trajectory deviations may lead to completely different execution results. For instance, for \emph{Open drawer} task, even if the end effector of the robot is only one centimeter away from the drawer handle, the drawer may not be successfully opened. Therefore, the learned representations should also be sensitive to interactions in diverse failure cases. 

To this end, we propose a hindsight interaction contrastive loss to improve interaction awareness by hindsight augmentation. 
Inspired by advances in Hindsight Experience Replay \cite{Andrychowicz2017HindsightER} and our variant works \cite{qian2022weakly,10155203} which find successful experiences from unsuccessful interactive data, the insight of our hindsight interaction contrastive loss is to obtain failure experiences from successful human demonstrations. 

As is shown in Fig. \ref{fig-aug}, hindsight augmentation consists of action augmentation and object augmentation. For each frame of one video $V^{h_i}$, action augmentation adds appropriate Gaussian noise to 3D body joint rotation $\theta_{body}$ obtained in Section \ref{section-3.1}, thereby obtaining the augmented 2D and 3D body keypoints after passing through the SMPL-X layer  \cite{pavlakos2019expressive}. Therefore, $z_{1:M}^{i,a\_aug}$ is obtained by inputting ${z_{action}^{i}}'$ and ${z_{obj}^{i}}$ to IAAformer. In addition, we use a simpler operation based on a certain probability, which uses action features extracted from other videos as ${z_{action}^{i}}'$. Similarly, object Augmentation replaces $z_{obj}^{i}$ of $V^i$ with $z_{obj}^{j}$ of $V^j$ to obtain $z_{1:M}^{i,o\_aug}$. To obtain the object features with $M$ timesteps, we only use the first frame of $z_{obj}^{j}$ and repeat $M$ times. 

After hindsight augmentation, our hindsight interaction contrastive loss is used to learn representations by minimizing the distance between positive pairs and maximizing the distance between negative samples. Specifically, we formulate our contrastive loss $L_{hic}$ for the positive pair $\{z_m^i,z_n^j\}$ as follows: 
\begin{equation}
\label{equ-loss2}
L_{hic}=-log\frac{exp(  sim(z_m^i ,  z_n^j) / \lambda_{temp} )}{sum_{neg}+sum_{a\_aug} +sum_{o\_aug}} 
\end{equation}
\begin{equation}
sum_{neg} = \sum_{k=1}^{N+M} \mathbbm{1}_{[k \ne m,n]} exp(sim(z_m^i ,  z_k^{i+j}) / \lambda_{temp})
\end{equation}
\begin{equation}
 sum_{a\_aug} = \sum_{k=1}^{N} exp(sim(z_m^i ,  z_k^{i,a\_aug}) / \lambda_{temp})
\end{equation}
\begin{equation}
sum_{o\_aug} = \sum_{k=1}^{N} exp(sim(z_m^i ,  z_k^{i,o\_aug}) / \lambda_{temp})
\end{equation}
where $sum_{neg}$ indicates the features in two sequences except the positive pair $\{z_m^i,z_n^j\}$ as negative samples, and $sum_{a\_aug}$ and $sum_{o\_aug}$ indicate the hindsight augmented features as negative samples.

In addition, we propose a temporal monotonic loss $L_{mon}$ to ensure the monotonicity of the matching. To allow gradient backpropagation, we define a differentiable matching index $MI(z_m^i, z^j_{1:N})$ for the nearest neighbor of $z_m^i$ in another feature sequence $z_{1:N}^j$ as follows:
\begin{equation}
MI(z_m^i, z^j_{1:N}) = \sum_{k=1}^{N}k \cdot \alpha_{m,k}
\label{equ-index}
\end{equation}
where $\alpha_{m,k}$ is defined in Equ. \ref{equ-3.7}. Then we formulate $L_{mon}$ as follows:
\begin{equation}
L_{mon} = \sigma(\frac{1}{M-1}\sum_{m=2}^{M} (MI(z_{m-1}^i, z^j_{1:N}) - MI(z_m^i, z^j_{1:N})))
\label{equ-loss3}
\end{equation}
where $\sigma$ is ReLU activation \cite{lecun2015deep}. 
Therefore, the total loss of IAAformer is the weighted sum of the three above losses, denoted as follows:
\begin{equation}
\label{equ-total-loss}
L_{total} = \gamma_{tcc}L_{tcc}+\gamma_{hic}L_{hic}+\gamma_{mon}L_{mon}
\end{equation}
where $\gamma_{tcc}=0.3$, $\gamma_{hic}=0.6$ and $\gamma_{mon}=0.01$, which are determined through the experiments described in Section \ref{sec-exp3}.


\subsection{Policy Learning by Inversion-Interaction}\label{section-3.4}

To imitate actions in human videos, 
we utilize a transformer-based Generative Adversarial Network \cite{goodfellow2014generative} to generate complex trajectories, denoted as TrajGAN. The network architecture is similar to \cite{10149371}, where four fully connected layers are added at the beginning.
We denote the trajectory extracted from one human video $V$ as $\tau \in \mathbf{R}^{steps \times 7}$, which contains 3-dimensional 3D keypoints $P_{3D}^{'}$, 3-dimensional rotation $\theta_{wrist}^{'}$ of the human wrist and one-dimensional gripper state (the contact state $c $). 
The generator $G$ of TrajGAN samples an 8-dimensional Gaussian noise $v \in \mathbf{R}^{8}$ as input and output a trajectory $\tau \in \mathbf{R}^{steps \times 7}$, which is fed into the discriminator $D$ to predict the probability of whether it comes from the true distribution. 
In addition, we add a contact discriminator $D_c$ to predict the true probability of the contact points of the generated trajectories. 


\begin{figure}[t]
	\centering
	\includegraphics[width=8cm]{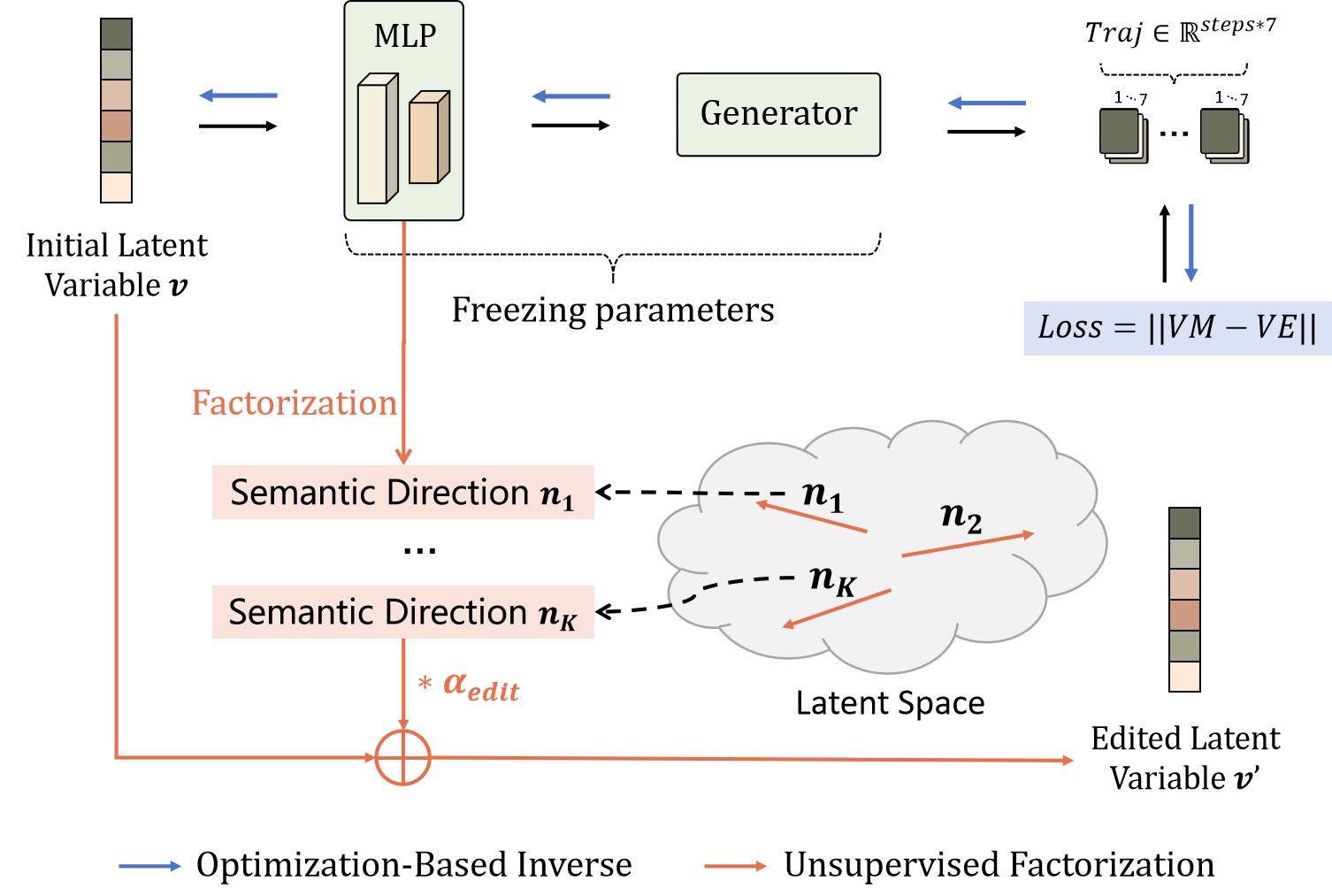}
	\caption{The schematic diagram of optimization-based GAN inversion for trajectory initialization and unsupervised semantic discovery for trajectory refinement.}
	\label{fig-method-edit}
	\vspace{-0.5cm}
\end{figure}

While TrajGAN has the potentials to generalize beyond the training data, the GAN-based architecture does not take environmental information as input. 
Therefore, TrajGAN needs to edit the generated trajectories to adapt to diverse scenario layouts.
We propose a two-stage \textbf{Inversion-Interaction} method,
which utilizes GAN inversion to optimize for coarse trajectories and further refine them through limited interaction. In addition, we propose an efficient sampling-based optimization method based on trajectory semantics (TS-CEM) to encourage safe exploration and improve sample efficiency. We summarize the whole algorithm of CIA in Algorithm \ref{algor1}.

\vspace{5pt}
\subsubsection{GAN Inversion for Fast Trajectory Initialization}\label{sec-3.4.1}
\

GAN inversion technique \cite{xia2022gan} is utilized to find the latent variable in the latent space of the pre-trained GAN with a generation instance.
We use optimization-based GAN inversion to obtain the latent variable $v$ by gradient backpropagation. 
While such methods in computer vision take an amount of iterative time to optimize from high-dimensional images, our TrajGAN
spend much less time in GAN inversion since the low dimensions of the output trajectories.

The schematic diagram of our GAN inversion is shown in Fig. \ref{fig-method-edit} through blue lines. Taken as input the latent variable $v$, a generator $G(\cdot)$ of the pre-trained TrajGAN outputs the trajectory $\tau \in \mathbf{R}^{steps \times 7}$, as $\tau=G(v)$. 
Since the optimal trajectory cannot be available, we utilize the position of the target object to constrain the contact points $C(\tau)$ in the generated trajectory.
We obtain the object position in test scenarios through RGBD camera calibration, which we denoted as the environmental variable $VE$. 
Therefore, we compute the optimization objective as follows:
\begin{equation}
\label{equ-3.4.3}
L_{inv}=\|VE-C(\tau)\|_2
\end{equation}
We set the completion of the GAN inversion when $\|VE-C(\tau)\|_2<=\eta$ ($\eta=0.02$) and optimization epochs $<1000$. 
Due to optimization on low-dimensional trajectories, the GAN inversion only requires about 20 seconds, and the convergence speed may not be affected by different initial latent variables.

Another use of GAN inversion is to measure the success confidence of tasks in testing scenarios, which can be used to reject the execution of tasks with particularly low confidence in advance. The confidence $\theta_{success}$ is computed as follows:
\begin{equation}
	\label{equ-confidence}
	\theta_{success}=\frac{2}{1+e^{L_{inv}} } 
\end{equation}
When $\theta_{success}<0.99$, it indicates that TrajGAN cannot generate appropriate trajectories. The reason here is that the position of the current environment variable $VE$ is too far from the distribution of the training data.

\vspace{5pt}
\subsubsection{Safe and Efficient Trajectory Refinement via Interaction}\label{sec-3.4.2}
\
The initialized trajectory still may not result in a high success rate in task execution, due to deviations between the contact position of the generated trajectory and the object, the inappropriate posture of the end effector, camera calibration errors, etc. Therefore, it is necessary to learn a policy to further refine the trajectory by interacting with the environment. 

The key issue in interacting with real environments is the trade-off between interaction safety and exploration efficiency. 
We assume that the initialized rough trajectory is reasonable, and therefore the interaction trajectory should not be too far from it to cause danger, nor too close to it to cause inefficient exploration.
To this end, we propose to discover semantic directions of trajectories in an unsupervised manner, which are utilized for reasonable and controllable editing of trajectories. 
Many works prove that GANs spontaneously have some directions in the latent space that represent semantic attributes of generations, such as the color, position and shape of objects \cite{qian2022weakly}.
In our case, the semantic direction may be the translation or rotation of the trajectory in a certain direction.
By correctly finding these semantic directions, we can use them to make meaningful edits to the generated trajectory and control the intensity of the edits.

\begin{algorithm}[t]
	\caption{Training Procedure for CIA}
	\label{algor1}
	\begin{algorithmic}[1]
		\State \textbf{Require:} Task videos: $V^h_{n=1:N}$, IAAformer $IAA$, TrajGAN $G_{\theta}$, Human Detection $OSX_{human}$, Object Detection $f_{obj}$, $M$ real world interactions per iteration, $Itr$ iteration numbers.
		\State $\%$ \textit{\textbf{Preprocessing Phase}}
		\State for n=1:N do
		\State \quad get learnable trajectories $\tau=OSX_{human}(V^h_n)$
		\State \quad get object image $I_{obj}=f_{obj}(V^h_n)$
		\State \quad Store all $\tau, I_{obj}$ into Data $D$
		\State end for
		\State $\%$ \textit{\textbf{Learning Phase}}
		\State Train $IAA$ with paired data by optimizing Equ. \ref{equ-total-loss}
		\State Train $G_{\theta}$ with trajectories $\tau$ sampled from $D$
		\State Get semantic directions $\{n_k\}_{k=1}^K$ by factoring the parameters $A$ in $\theta$
		\State $\%$ \textit{\textbf{Adaptation Phase}}
		\State $\%$ \textit{\textbf{Inversion}}
		\State Initialize the latent variable $v_{init}$ by optimizing Equ. \ref{equ-3.4.3}
		\State $\%$ \textit{\textbf{Interaction}}
		\State for i=1:Itr do 
		\State \quad for m=1:M do
		\State \quad \quad Sample a semantic direction $n_k$ from $k \sim N\{\mu_k,\sigma_k\}$ and an editing scale $\beta$ from $N(\mu_\beta,\sigma_{\beta})$
		\State \quad \quad Execute $\tau_{new}=G_{\theta}(v_{init}+\beta n_k)$, collect video: $V^r_m$
		\State \quad end for
		\State \quad Compute and rank the reward by Equ. \ref{equ-reward} for every $m$
		\State \quad Pick \{elite trajectories\} to fit $\mu_k,\sigma_k,\mu_\beta,\sigma_{\beta}$
		\State end for
		\State \textbf{return} $\tau_{final}=G_{\theta}(v_{init}+\beta n_k)$
		
	\end{algorithmic}
\end{algorithm}

In TrajGAN, each layer learns a transformation from input to output. Focusing on the first four fully connected layers $G_{fc}$, the affine transformation on the latent variables $v \in \mathbb{R}^{8}$ can be expressed as follows:
\begin{equation}
\label{equ-3.4.1}
G_{fc}(v) \triangleq y = Av + b
\end{equation}
where $y \in \mathbb{R}^{C_y}$ is the $C_y$-dimensional feature. $A \in \mathbb{R}^{C_y \times 8}$ and $b \in \mathbb{R}^{C_y}$ denote the weight and bias of $G_{fc}$ respectively.
Like prior works \cite{qian2022weakly, shen2021closed}, 
the editing model on the latent variables is formulated as follows:
%
\begin{equation}
\begin{aligned}
\label{equ-3.4.3}
y' \triangleq G_{fc}(v')&=G_{fc}(v+\beta n)=y + \beta An
\end{aligned}
\end{equation}
where $\beta$ denotes the editing scale.
We can see that given any latent variable $v$, we can utilize $\beta An$ to adjust the intensity of editing. 
Therefore, the weight parameter $A$ should contain rich knowledge about the semantic attributes of the trajectory. 
The problem of discovering semantic directions can be factorized by solving the following optimization problem:
\begin{equation}
\begin{aligned}
\label{equ-3.4.4}
N^* = \mathop {argmax} \limits_{N \in \mathbb{R}^{8 \times K}: n_k^Tn_k=1 }\|An_k\|^2_2
\end{aligned}
\end{equation}
where $N^*=\{n_1,n_2,...,n_K \}$ denotes the top-$K$ important semantics. Lagrange multipliers ${\{ \lambda_k\}^{K}_{k=1}}$ are introduced to solve this problem as follows:
\begin{equation}
\begin{aligned}
\label{equ-3.4.5}
N^* &= \mathop {argmax} \limits_{N \in \mathbb{R}^{8 \times K}}\|An_k\|^2_2 - \sum_{k=1}^{K} \lambda_{k}(n_k^Tn_k-1)\\
& = \mathop {argmax} \limits_{N \in \mathbb{R}^{8 \times K}} \sum_{k=1}^{K} (n_k^TA^T A n_k - \lambda_{k}n_k^Tn_k + \lambda_{k})
\end{aligned}
\end{equation}
By taking the partial derivative of $n_k$ in Equ. \ref{equ-3.4.5}, we obtain
\begin{equation}
\label{equ-3.4.6}
2A^T A n_k - 2\lambda_{k}n_k=0
\end{equation}
Therefore, we can calculate the eigenvectors of the matrix $A^TA$ and select the eigenvectors with the $K$ largest eigenvalues as the semantic directions $\{n_k\}_{k=1}^K$. With several proper semantic directions, agents can edit trajectories in a controllable and reasonable manner to explore in the environment.

\vspace{5pt}
\subsubsection{Optimization Procedure Based on Trajectory Semantics}\label{sec-3.4.3}
\

Reinforcement learning algorithms suffer from low sampling efficiency in vast action spaces and complex environments.
To this end, we introduce Cross Entropy Method \cite{de2005tutorial} based on Trajectory Semantic sampling (TS-CEM) as an efficient gradient-free optimization method.

Given the task and the initial latent variable $v_{init}$, we sample a semantic direction $n_k$ from $k \sim N\{\mu_k,\sigma_k\}$ and an editing scale $\beta$ from $N(\mu_\beta,\sigma_\beta)$. And we obtain the corresponding trajectory $\tau_{new}=G_{\theta}(v_{init}+\beta n_k)$. 
Therefore, we execute R=20 trajectories in the real world and record the corresponding videos $V^r$. To iteratively improve the policy without human supervision, we utilize IAAformer to compute the alignment distances of randomly sampled human demonstrations $V^h$ and robot execution sequences $V^r$ as reward functions. The reward is computed as follows:
\begin{equation}
	\label{equ-reward}
	Rew = \sum_{t=0}^{T}||z^{r}_t - \hat{z}^{h}_t||_2^2
\end{equation}
where $z^r_t$ denotes the frame feature in the robot's execution video and $\hat{z}^h_t$ denotes the corresponding nearest neighbor frame in a randomly selected human demonstration video.
Based on such rewards, we rank these trajectories and consider the trajectories of the top five highest rewards as elite trajectories, which can be used to compute the distribution parameters $\mu_k, \sigma_{k}, \mu_\beta, \sigma_\beta$ of the semantic direction and the editing scale in the next iteration.


\section{Experimental result and analysis}\label{section4}
Various experiments are conducted to validate the effectiveness of the framework CIA and each component.
We first introduce our experimental setup in Section \ref{sec-4.1}. In the following sections, our experiments tend to address these questions:
\begin{enumerate}
	\item How does our CIA compare with prior state-of-the-art works in imitation learning from videos?
	\item Does IAAformer match two videos well over time and understand the interaction process?
	\item What is the effect of the initialization based on GAN inversion and the editing based on trajectory semantics in the TrajGAN?
	\item Is each component of CIA really effective for learning from human videos?
	\item How well can CIA adapt to novel scenario layouts, and novel object instances?
\end{enumerate}

\subsection{Experimental Setup}\label{sec-4.1}

\subsubsection{Hardware Platform}\label{sec-4.1.1}
\

\begin{figure}[t]
	
	\centering
	\subfloat[The robot platform and six tasks in our experiments.]{\small
		\begin{minipage}{8.cm} 
			\label{exp-task-pic-a}
			\includegraphics[width=\textwidth]{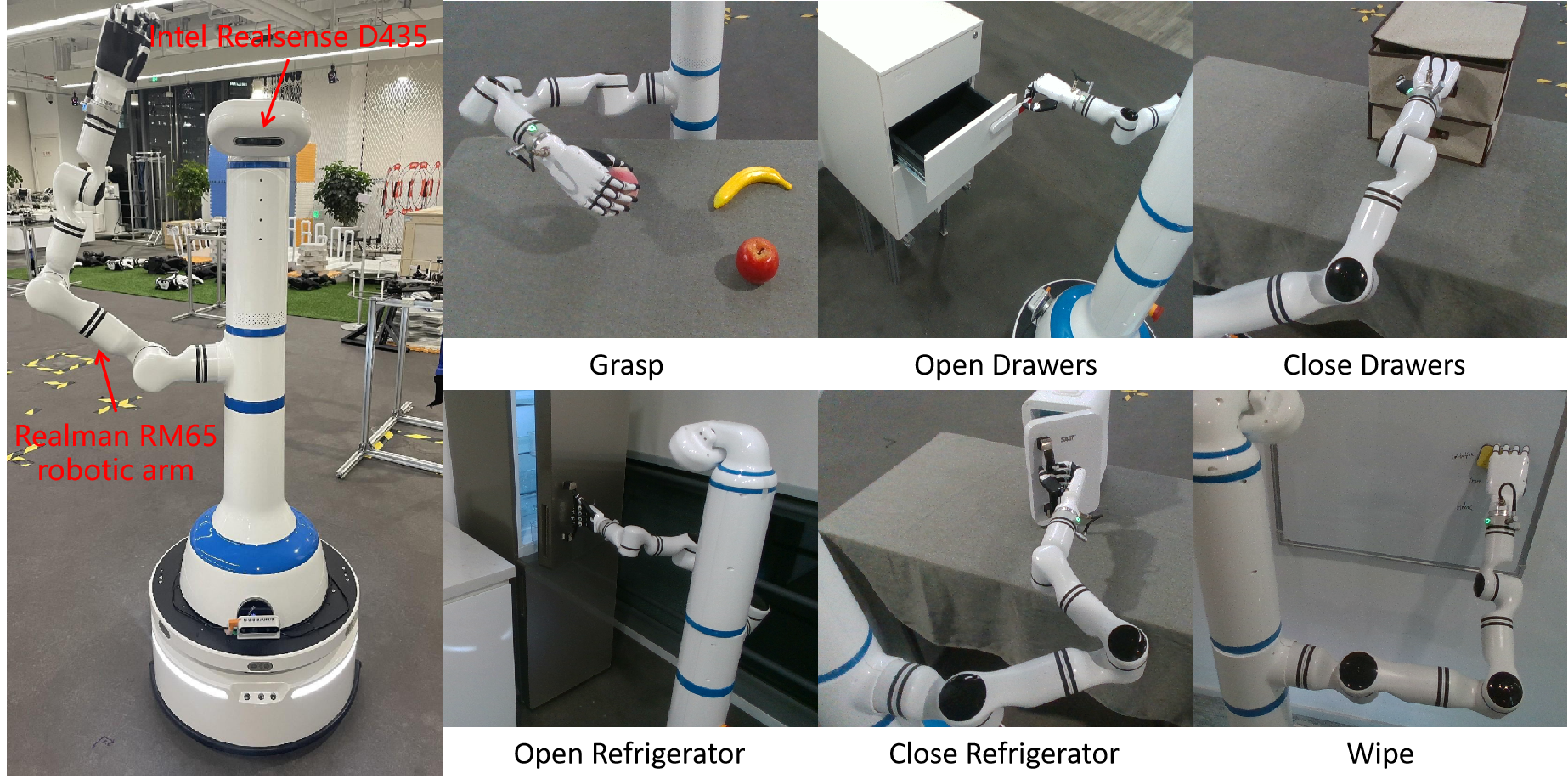} \\
		\end{minipage}
	}
	
	\subfloat[Examples of objects used for experimentation.]{
		\begin{minipage}{8.cm}
			\label{exp-task-pic-b}
			\includegraphics[width=\textwidth]{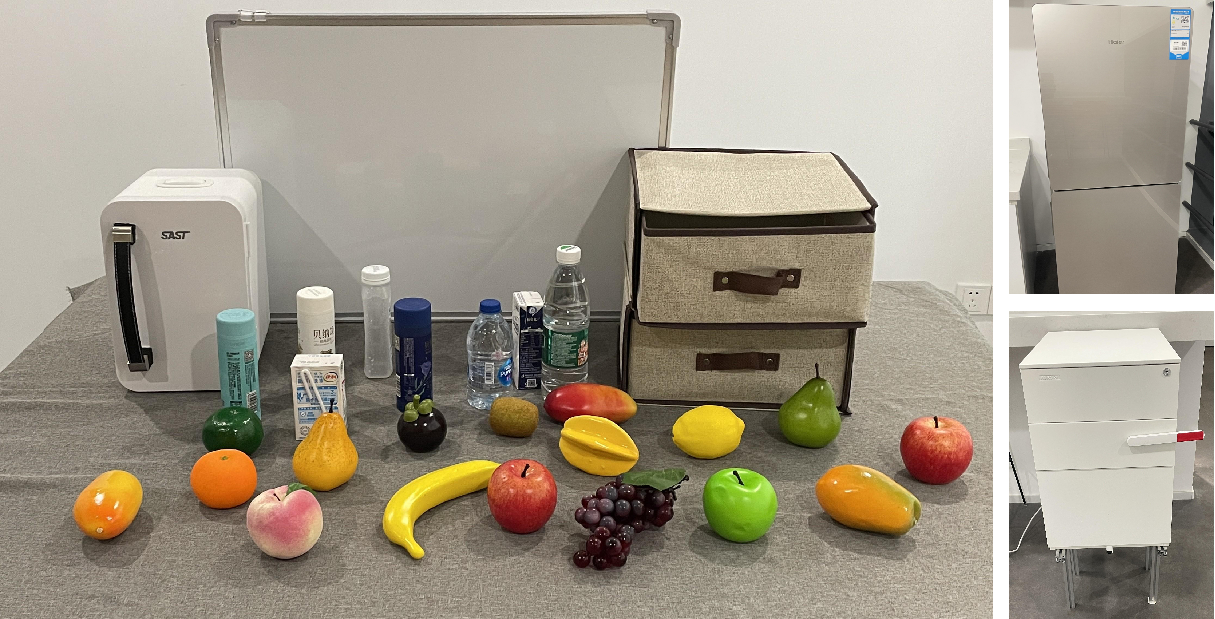} \\
			
		\end{minipage}
	}

	\caption{Illustration of experimental settings in the real world.} 
	\label{exp-task-pic-all}

\end{figure}

To perform experiments in various scenarios, we use a Realman RM65 robotic arm mounted horizontally on a mobile chassis with an Inspire five-finger dexterous hand as our humanoid agent, as is shown in Fig. \ref{exp-task-pic-all}.
We use a Cartesian position controller to translate the end effector of the robotic arm and an in-built orientation controller to control the rotation.
We simplify the degree of freedom of the hand into a 1-dimensional binary on-off state. An Intel Realsense D435 camera is utilized, which is mounted either on the robot head or externally fixed, depending on the task. In addition, we use 6 NVIDIA V100 GPUs to train all the networks. The full training takes about 5 hours.

\subsubsection{Manipulation Tasks and Training Data}\label{sec-4.1.2}
\

To validate our framework CIA, we select six tasks that can be abundantly found in existing human daily behavior video datasets.
The tasks include \emph{Grasping}, \emph{Opening refrigerator}, \emph{Closing refrigerator}, \emph{Opening drawer}, \emph{Closing drawer}, and \emph{Wiping} shown in Fig. \ref{exp-task-pic-a}. Note that the human videos of these tasks lack corresponding robot actions, and they vary in terms of viewpoints, execution embodiments, and scenario layouts. The objects used in the testing scenarios are shown in Fig. \ref{exp-task-pic-b}. 

We choose 2853 videos in Dex-YCB \cite{chao2021dexycb} for \emph{Grasping} of diverse objects in diverse viewpoints. 
Each other task has 365-950 videos from VLOG \cite{fouhey2018lifestyle} and Charades \cite{sigurdsson2016hollywood}, where various people record these everyday tasks with diverse objects in various environments.
To select appropriate videos, we compute the temporal alignment distance between each video of the same task and the other 50 videos, and eliminate videos whose average temporal alignment distance is more than the threshold 0.5.

\vspace{5pt}
\subsubsection{Evaluation Metric}\label{sec-4.1.3}
\

\begin{table}[t]
	\setlength{\tabcolsep}{0.3em}
	
	\caption{Comparison experiments with other baselines on six real-world tasks}
	\label{table-cia}
	\centering
	
	\footnotesize{
		\begin{tabular}{m{1.58cm}<{\raggedright}| m{1cm}<{\centering}| m{1cm}<{\centering}|m{1.15cm}<{\centering}|m{1cm}<{\centering}|m{1cm}<{\centering}|m{1.cm}<{\centering}}
			\Xhline{0.5pt}
			\toprule
			\multirow{4}{*}{\textbf{Method}}           &  \multicolumn {6}{c}{\textbf{Success Rate (\%)} }\\ 
			\cmidrule(lr){2-7}
			& \multirow{2}{*}{\textbf{Grasp}} &  \textbf{Open}  & \textbf{Open}  &\multirow{2}{*}{\textbf{Wipe}} & \textbf{Close}& \textbf{Close} \\
			&  &  \textbf{Drawer}  & \textbf{Refrig.}  & & \textbf{Drawer}& \textbf{Refrig.} \\
			\midrule
			BC 		 						&0.40 & 0.33 & 0.23 & 0.53&0.47 &0.40 \\
			CQL-CLS  						&0.37 & 0.20 &  0.17 & 0.47 &0.33 &0.37 \\
			CQL-CASA 						& 0.47& 0.27 &  0.20 &  0.47 &0.50 & 0.43\\
			
			CQL-ours 						& 0.57 &0.40 &0.27 &0.57&0.57 & 0.53\\
			\midrule
			CIA-w/o-Iter				&0.43 &0.37 &0.33 &	0.50 &0.67 &0.63 \\
			CIA-w/-1Iter				&0.77 &0.63 &0.57 & 0.73&0.83 & 0.87\\
			\textbf{CIA-w/-2Iter} 	& \textbf{0.90}& \textbf{0.87} & \textbf{0.73} & \textbf{0.93}& \textbf{0.97}& \textbf{0.93}\\
			
			\midrule
			Trial number & 30 & 30 & 30 & 30 & 30 & 30\\
			\bottomrule
			\Xhline{0.5pt}
		\end{tabular}
	}
	\vspace{-1em}
\end{table}

We use the success rate to measure the performance of the methods. Following the temporal alignment literature \cite{dwibedi2019temporal, haresh2021learning, kwon2022context}, we utilize three metrics for the evaluation of our IAAformer. After training the IAAformer, we freeze the parameters and evaluate in the trained representation space.

\vspace{2pt}
\noindent\textbf{Success Rate} is utilized to measure the performance of CIA and other baselines. Success is determined when the robot's execution video matches with the human demonstration.

\vspace{2pt}
\noindent\textbf{Phase Classification} evaluates the representation quality based on the per-frame fine-grained classification accuracy. We train an SVM classifier on an annotated subset of the training dataset to predict phase categories.

\vspace{2pt}
\noindent\textbf{Phase Progression} measures the ability of representations to capture the progress of an action over time. A linear regressor is used to predict the phase progression values. It is computed as the average $R$-squared measure between the predictions and the ground truth progress values.

\vspace{2pt}
\noindent\textbf{Kendall’s Tau} is a statistical measure to measure the monotonicity of matching order over time. Compared to the above two metrics, it does not require additional labels. It is in the range of $[-1,1]$, where 1 denotes perfect alignments and -1 denotes alignments in reverse order.

\subsection{Evaluation of CIA and Comparison to Baselines}

To answer the first question, we compare CIA to several state-of-the-art baselines. 
Since performing general RL in the real world is time-consuming, we compare with a SOTA offline RL, Conservative Q-learning (CQL) \cite{kumar2020conservative}, which is trained on offline datasets without interaction. We follow \cite{shao2021concept2robot, nair2022learning} to train a classifier to determine whether human videos and robot videos belong to the same task category as the sparse reward. We call this baseline \textbf{CQL-CLS}. And we compare against a competing SOTA method for learning representations by temporal alignment. We refer to \textbf{CQL-CASA} to compute the distance in the CASA representation space as the reward. We also train CQL using rewards computed by our IAAformer, denoted as \textbf{CQL-ours}. In addition, we compare to behavior cloning (\textbf{BC}), which predicts the actions from the input, reflecting prior works \cite{qian2023robot,zhang2022one}. 
We found that these baselines hardly predict correct actions from raw pixel input, due to the significant visual differences between human videos and test scenarios. Therefore, we use the three-dimensional contact points as the object's coordinates, along with the robotic end-effector position, which are fed into 3 fully connected layers to predict seven-dimensional actions described in Section \ref{sec-4.1.2}.

%
The comparison results on six real-world everyday tasks are shown in Table. \ref{table-cia}. Success rates are computed out of 30 trials and each task contains different object categories and locations. We also show the success rates of our CIA without and with 1 or 2 iterations.
Note that in each iteration, the robot executes 20 trajectories, and each trajectory takes approximately 40 seconds. So each iteration takes about 15 minutes.
As is shown, our CIA significantly outperforms all the baselines. In particular, offline RL methods are difficult to complete tasks and have poor generalization performances. Interestingly, CQL-ours outperforms CQL-CLS and CQL-CASA, which indicates that our objective function can identify good interaction trajectories and task progress. The Behavior Cloning method performs similarly to CQL-CASA in that optimization is performed by averaging the error at each step, which results in inaccuracies at critical contact points with the object.
\begin{figure}[t]
	\centering
	\includegraphics[width=8.8cm]{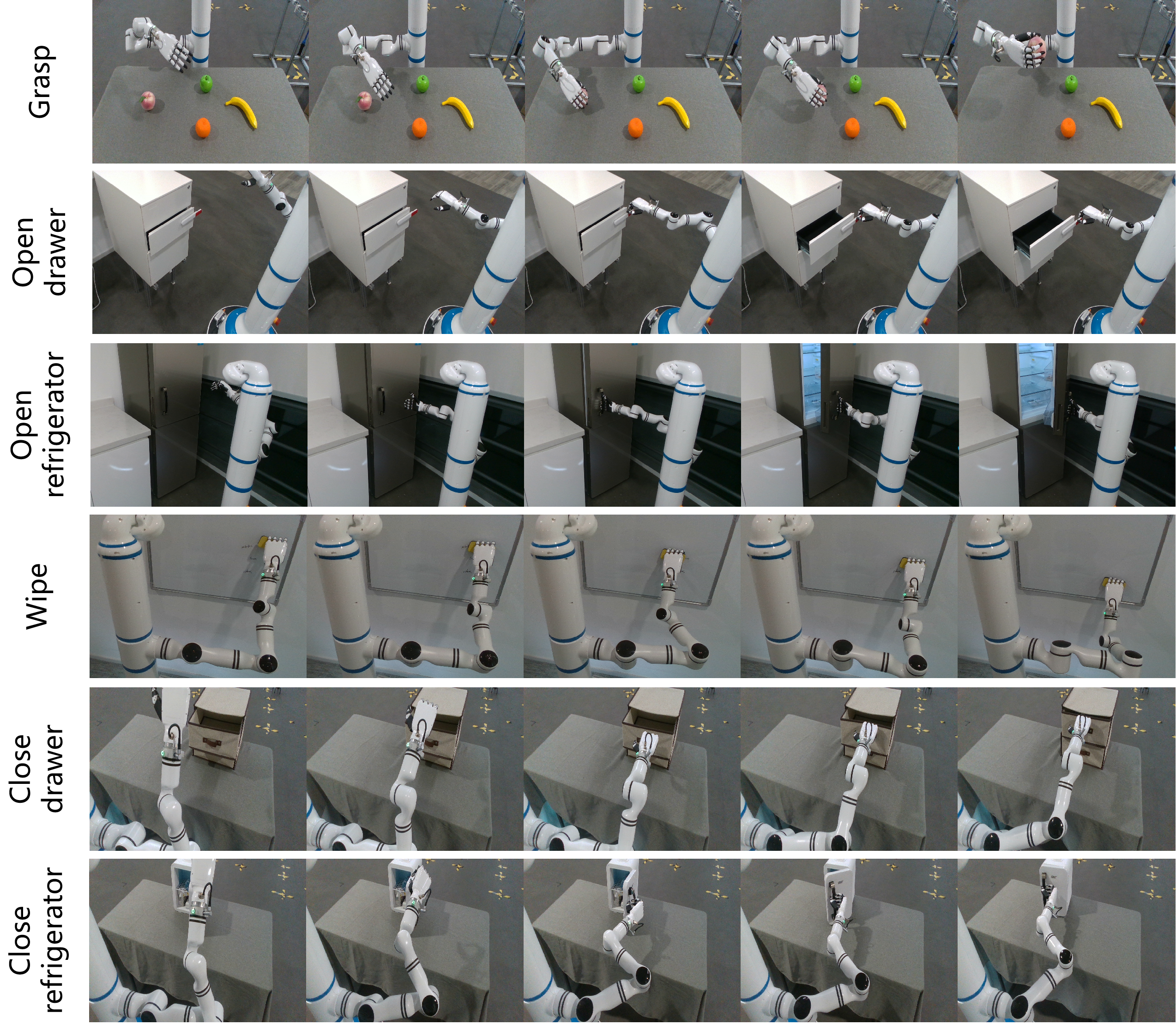}
	\caption{Examples of successful trajectories for 6 tasks.}
	\label{exp-success}
	\vspace{-0.5cm}
\end{figure}
As is shown, our CIA can improve the performance iteratively, and CIA with two iterations can achieve a success rate of 90\% on most tasks. Some examples of successful execution trajectories on six tasks are shown in Fig. \ref{exp-success}. However, we note that the success rates of \emph{Closing drawer} and \emph{Closing refrigerator} tasks are lower than other tasks since these tasks require high-precision movements.
In particular, the robot is also limited by its arm span when opening a large refrigerator.

\subsection{Comparison With Prior Temporal Alignment Methods}\label{sec-exp3}


To answer the second question, we compare IAAformer against previous state-of-the-art temporal alignment methods. \textbf{TCC} \cite{dwibedi2019temporal} proposes a cycle-consistency loss to learn the visual features. 
\textbf{LAV} \cite{haresh2021learning} proposes a Soft-DTW loss and a temporal regularization terms to learn frame-level features. 
\textbf{CASA} \cite{kwon2022context} utilizes a contrastive loss to temporally align the pair of 3D skelatal sequences. 
For a fair comparison, we modify the output feature dimensions of these feature extractors to be the same as our IAAformer.

\begin{figure}[t]
	\centering
	\includegraphics[width=8.8cm]{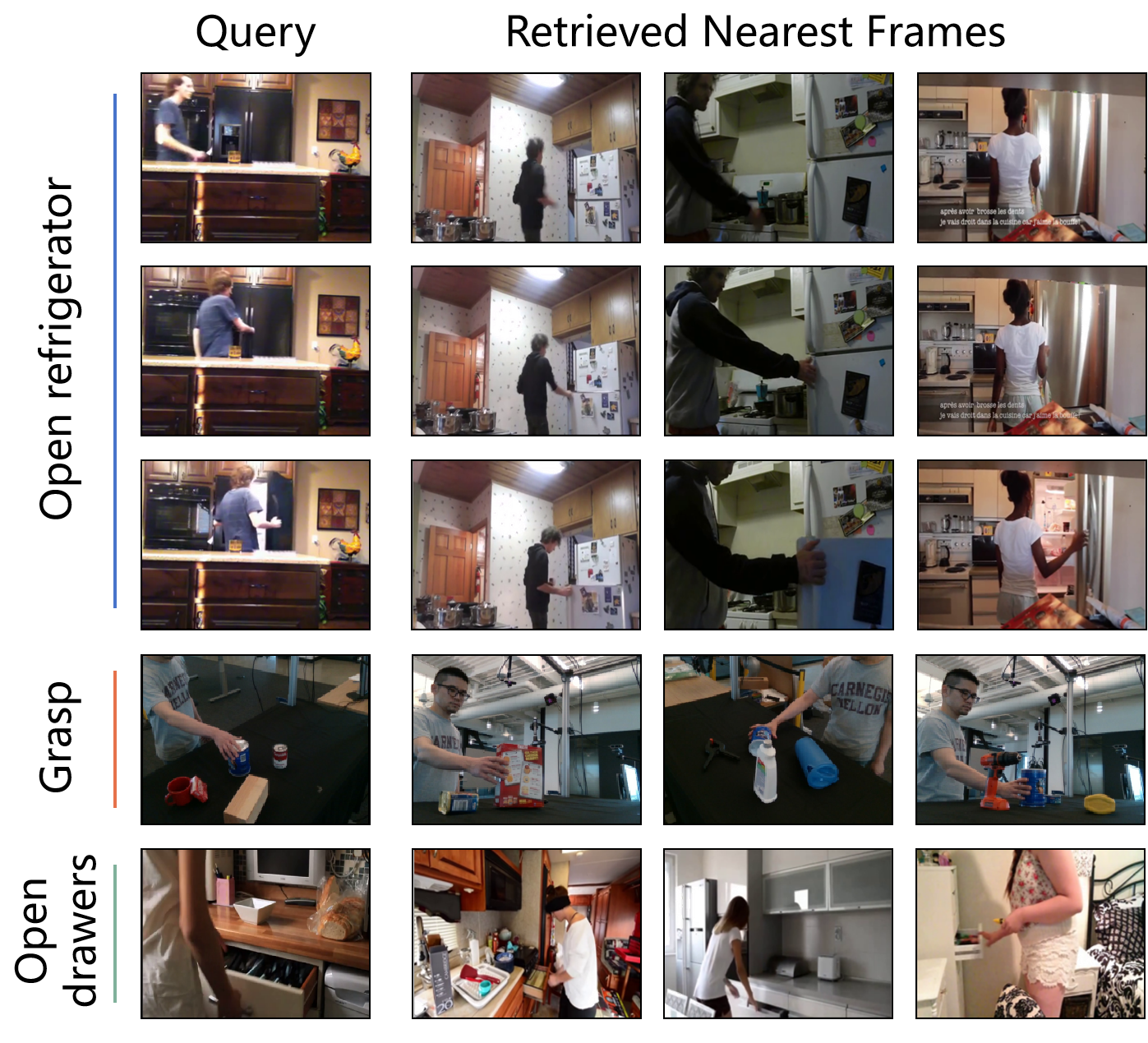}
	\caption{Some fine-grained retrieval results of different tasks based on the query frames. 
	}
	\label{exp-align-pic}
	\vspace{-1.5em}
\end{figure}

\begin{table}[t]
	\setlength{\tabcolsep}{0.3em}
	
	\caption{Comparison experiments with other methods on temporal alignment}
	\label{table-align-1}
	\centering

	\footnotesize{
		\begin{tabular}{c|c|c|c|c|c}
			\Xhline{0.5pt}
			\toprule
			&                & \textbf{Input} &  \textbf{Phase}  & \textbf{Phase}  & \textbf{Kendall's}\\ 
			\textbf{Task}&\textbf{Method} & \textbf{Type} &  \textbf{Class(\%)}  & \textbf{Progression}  & \textbf{Tau} \\
			\midrule
			\multirow{4}{*}{Refrig}&TCC\cite{dwibedi2019temporal} &Image & 36.61 & 0.4358 & 0.4813\\
			&LAV\cite{haresh2021learning} &Image & 41.89 &  0.5833 & 0.6348  \\
			&CASA\cite{kwon2022context} & 3D Pose& 63.07 &  0.6808 & 0.7606  \\
			
			&\textbf{IAA(ours)} & \textbf{3D Pose+Obj}& \textbf{78.94} & \textbf{0.8351} & \textbf{0.8502}\\
			
			\midrule
			\midrule
			\multirow{4}{*}{Grasp}&TCC\cite{dwibedi2019temporal} &Image & 76.79 & 0.5513 & 0.3859\\
			&LAV\cite{haresh2021learning} &Image & 81.71 &  0.4955 & 0.5177  \\
			&CASA\cite{kwon2022context} & 3D Pose& 85.10 &  0.6125 & 0.6909  \\
			
			&\textbf{IAA(ours)} & \textbf{3D Pose+Obj}& \textbf{93.65} & \textbf{0.9027} & \textbf{0.9341}\\
			
			\bottomrule
			\Xhline{0.5pt}
		\end{tabular}
		
	}
	\vspace{-0.5em}
\end{table}

\begin{table}[t]
	\setlength{\tabcolsep}{0.3em}
	
	\caption{Comparison experiments with different loss coefficients of IAAformer on \emph{Open refrigerator} task}
	\label{table-align-22}
	\centering

	\footnotesize{
		\begin{tabular}{c|c|c|c|c|c|c}
			\Xhline{0.5pt}
			\toprule
			& & & &  \textbf{Phase}  & \textbf{Phase}  & \textbf{Kendall's} \\ 
			\textbf{Model}& $\mathbf{\gamma_{tcc}}$ & $\mathbf{\gamma_{hic}}$ & $\mathbf{\gamma_{mon}}$&   \textbf{Class(\%)}  & \textbf{Progression}  & \textbf{Tau} \\
			\midrule
			1& 0.6 & 0.3 & 0.01 & 77.61 &  0.8266 & 0.8451  \\
			2& 0.5 &0.5 & 0.01 & 77.92 & 0.8309  & 0.8447  \\
			3&\textbf{0.3} & \textbf{0.6}& \textbf{0.01} & \textbf{78.94} & \textbf{0.8351} & \textbf{0.8502}\\
			
%
			
			\bottomrule
			\Xhline{0.5pt}
		\end{tabular}
		
	}
	\vspace{-1em}
\end{table}

The comparison experiment results of \emph{Opening refrigerators} and \emph{Grasping} tasks are presented in Table. \ref{table-align-1}. IAAformer significantly outperforms other methods in three evaluation metrics for two task datasets. For TCC \cite{dwibedi2019temporal} and LAV \cite{haresh2021learning}, the entire image input resulted in lower performance due to the lack of explicit structured information (e.g. hand trajectories, object coordinates of interest). CASA \cite{kwon2022context} shows some improvements, but the improvements are limited due to the lack of object coordinates which provide interactive cues. In contrast, our IAAformer explicitly accounts for hand and object information through self- and cross-attention mechanisms, thereby improving the understanding of interactive processes. Some fine-grained retrieval results of different tasks are visualized in Fig. \ref{exp-align-pic}. It shows that IAAformer can successfully retrieve temporal similar frames from other videos based on one query frame.

To study the impact of different loss coefficients on alignment performance, we present the results of comparative experiments on \emph{Open refrigerator} task in Table. \ref{table-align-22}. As is shown, model 3 with $\gamma_{tcc}=0.3$, $\gamma_{hic}=0.6$ and $\gamma_{mon}=0.01$ has slightly better performance on the three evaluation metrics, which demonstrates the effectiveness of our hindsight interaction contrastive loss.

\begin{figure}[t]
	\centering
	\includegraphics[width=8.8cm]{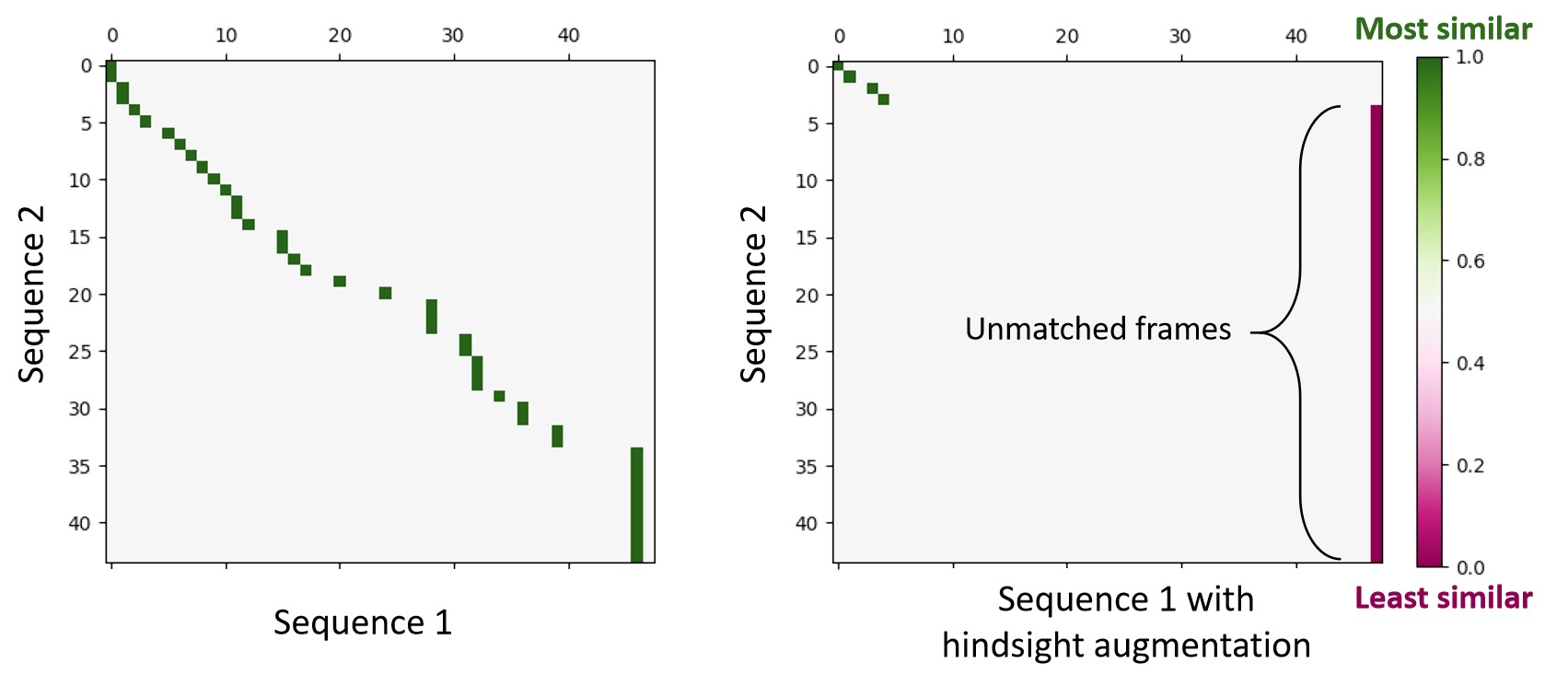}
	\caption{Alignment results between two videos without and with hindsight augmentation. For each frame in sequence 2, we find the frame with the highest similarity in sequence 1. 
		}
	\label{fig-exp-aug-align}
	\vspace{-1.5em}
\end{figure}

\noindent\textbf{Interaction awareness of IAAformer.} To demonstrate IAAformer's understanding of the interaction process, we visualize the alignment results between two sequences without and with the hindsight augmentation, which are shown in Fig. \ref{fig-exp-aug-align}. The left subfigure shows that two original videos with different time lengths match well. For each frame in sequence 2, we find the frame with the highest similarity in sequence 1. For clarity, we set the value of the most similar frame to 1 and the others, which are below the threshold of 0.3, to 0. In the right subfigure, we perform minor action augmentations to sequence 1, resulting in an unsuccessful interaction. It can be seen that IAAformer can distinguish the mismatch between two interaction processes after the first few frames. This indicates that using IAAformer can be used to compute the reward function for subsequent policy improvement.

\subsection{Trajectory Initialization and Controllably Editing}

\begin{figure}[t]
	\centering
	\includegraphics[width=8cm]{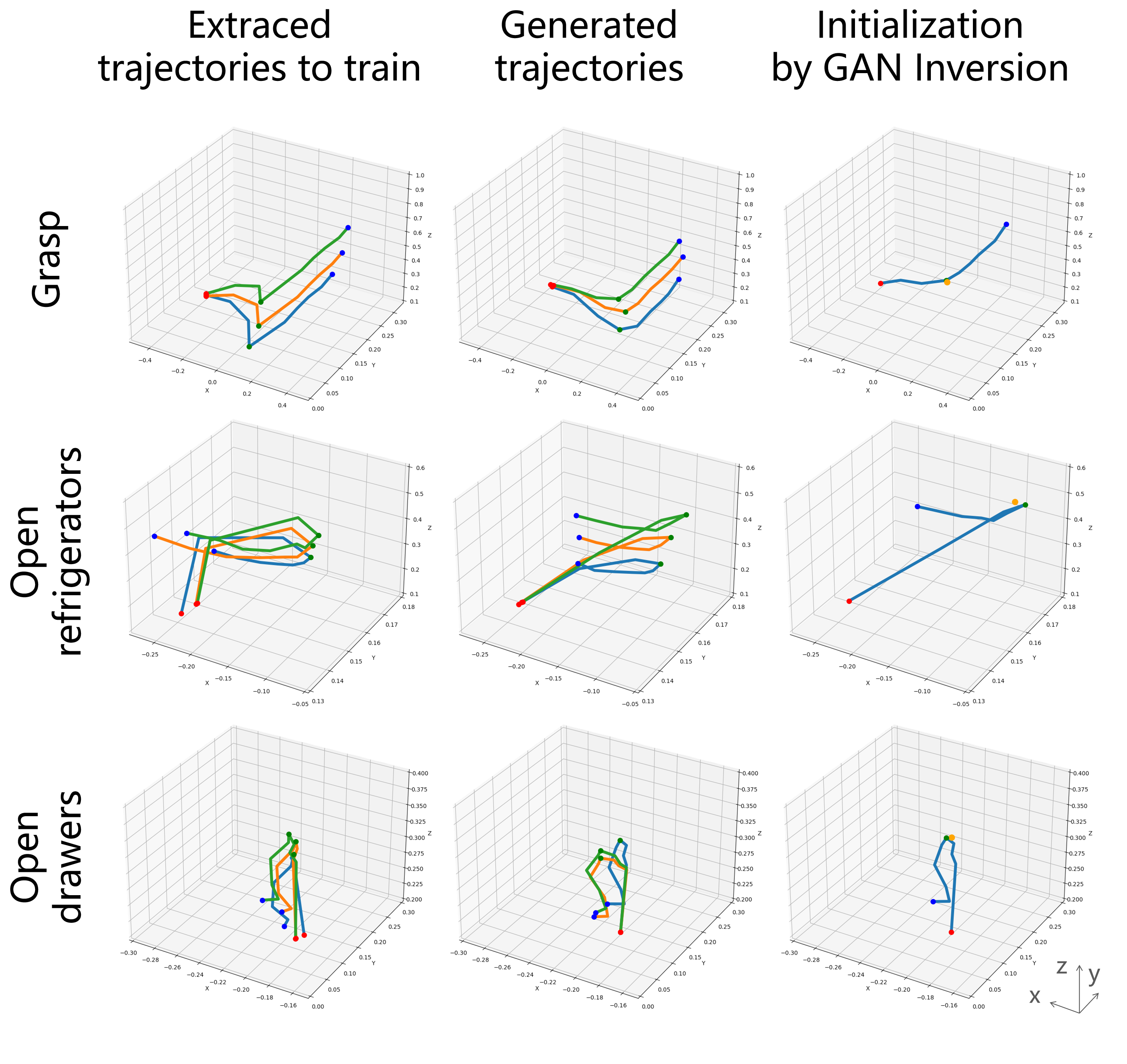}
	\caption{Examples of generating trajectories of some tasks and trajectories initialized based on GAN Inversion. The red, green, blue and orange points represent the starting point, the contact point, the endpoint and the environmental variable respectively.
	}
	\label{fig-exp-generate-traj}
	\vspace{-0.5cm}
\end{figure}

To answer the third question, we demonstrate the results of task-specific TrajGANs for three tasks: \emph{Grasp}, \emph{Open refrigerator} and \emph{Open drawer} in Fig. \ref{fig-exp-generate-traj}. 
A small number of extracted trajectories and generated trajectory examples are shown in the left two columns while the third column shows the trajectories initialized by GAN inversion.
We let the red, green, and blue points represent the starting point, the contact point, and the endpoint, respectively. 
In addition, the orange point represents the environmental variable, which is the expected contact point. 
As is shown, GAN inversion can be utilized to initial trajectories well by minimizing the distance between the environmental variable and the contact point.

\begin{figure}[t]
	\centering
	\includegraphics[width=8.8cm]{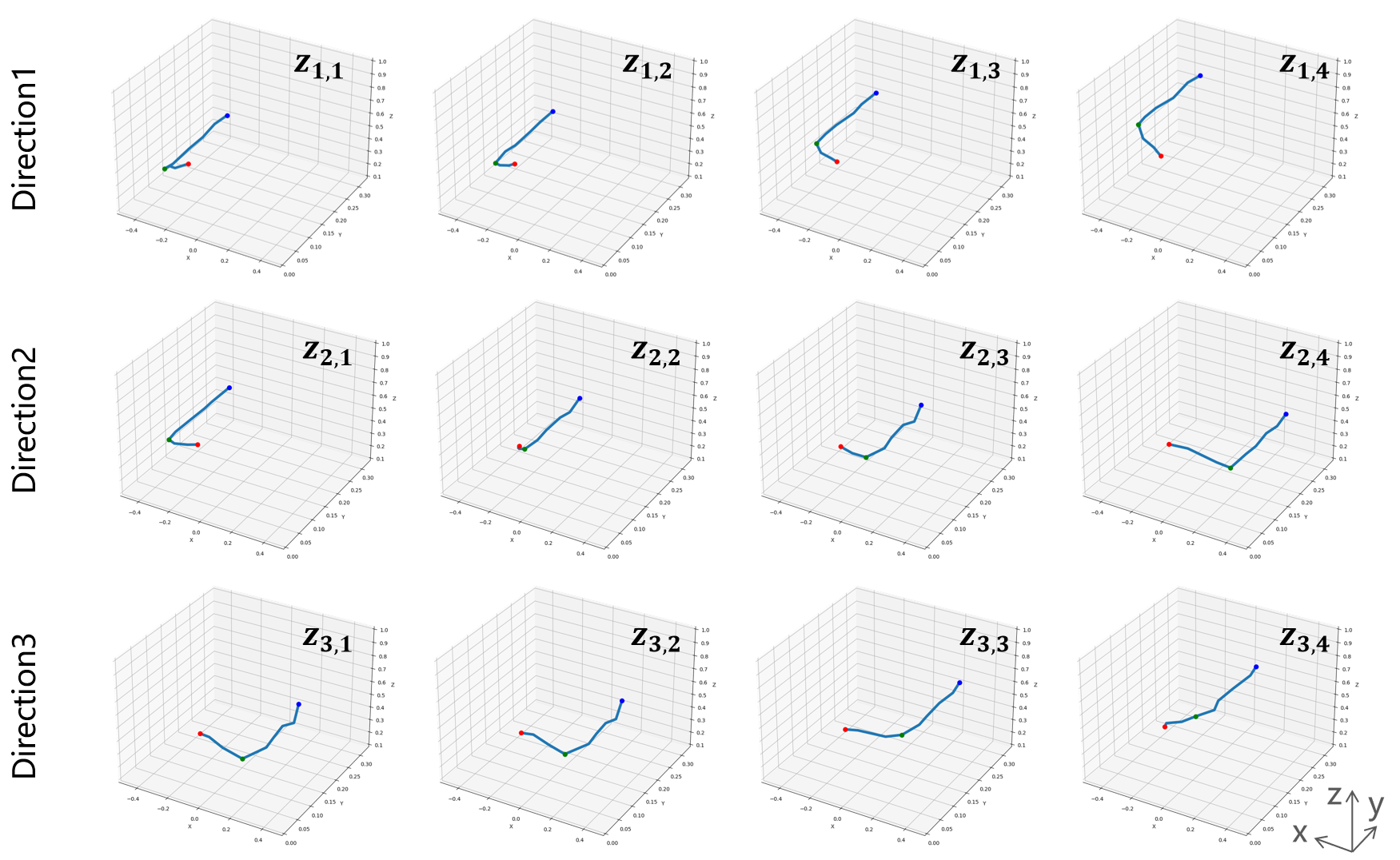}
	\caption{The schematic diagram of trajectories based on semantic direction editing on the \emph{Grasp} task.}
	\label{exp-direction}
\end{figure}

Controllable editing capabilities based on the semantic directions are crucial for agents to safely explore the environment. To demonstrate the semantic directions by unsupervised discovery, we edit the latent codes in the TrajGAN latent space of the \emph{Grasp} task, which are visualized in Fig. \ref{exp-direction}. 
Three main semantic directions are discovered by the factorization of parameters described in Method \ref{sec-3.4.2}. 
Specifically, taking the first line as an example, we randomly sample a latent variable $z_{1,1}$ and gradually add a weighted direction $n_1$. Then, TrajGAN takes as input the edited latent variable $z_{1,k}=z_{1,1}+\beta_i n_1, i=2,3,4$ and generates the trajectory $v_{1,k}$ corresponding to the semantic changes. 
From left to right, as the editing scale $\beta_i$ increases, the contact points of the generated trajectories move towards the z-axis. 
Therefore, by controlling the editing scale and semantic directions, small adjustments can be made near the initialization trajectory, which can be used for safe exploration in real environments.


\subsection{Ablation Analysis of Each Component of CIA}
To answer the fourth question about the effectiveness of each component, we first perform several ablation experiments to analyze each module of our IAAformer on \emph{Open refrigerator} and \emph{Grasp} tasks, which is shown in Table. \ref{table-ablation-1}. For simplicity, we use IAA to represent IAAformer. These modules include hindsight interaction contrastive loss, temporal monotonic loss and attention mechanisms, referred to as HA, ML and Atten respectively. As is shown, IAA with 3D pose as input can perform poorly like CASA, which indicates the importance of object information as input for understanding task processes. We found that IAA w/o HA performs well in Kendall’s Tau while IAA w/o ML performs well in the first two metrics. It demonstrates that the temporal monotonic loss can help improve the alignment monotonicity and the hindsight interaction contrastive loss can help improve the ability of representation to understand the task process. In addition, self- and cross-attention mechanisms can further improve the temporal alignment performance of IAAformer.

\begin{table}[t]
	\setlength{\tabcolsep}{0.3em}
	
	\caption{Ablation for each component of our IAAformer}
	\vspace{-0.1cm}
	\label{table-ablation-1}
	\centering

	\footnotesize{
		\begin{tabular}{c|c|c|c|c|c}
			\Xhline{0.5pt}
			\toprule
			\multirow{2}{*}{\textbf{Task}}&  \multirow{2}{*}{\textbf{Method}}             & \textbf{Input} &  \textbf{Phase}  & \textbf{Phase}  & \textbf{Kendall's}\\ 
			& & \textbf{Type} &  \textbf{Class(\%)}  & \textbf{Progression}  & \textbf{Tau} \\
			\midrule
			\multirow{5}{*}{Refrig.}&IAA & 3D Pose &69.72 &0.7194 &0.7463\\
			&IAA w/o HA & 3D Pose+Obj &73.14 &0.7633 & 0.8107 \\
			&IAA w/o ML & 3D Pose+Obj &74.95 &0.7879 &0.7562\\
			&IAA w/o Atten & 3D Pose+Obj &78.16 &0.8292 &0.8385 \\
			&\textbf{IAA(ours)} & \textbf{3D Pose+Obj}& \textbf{78.94} & \textbf{0.8351} & \textbf{0.8502}\\
			
			\midrule
			\midrule
			\multirow{5}{*}{Grasp}&IAA & 3D Pose &85.82&0.7583&0.7794\\
			&IAA w/o HA & 3D Pose+Obj &86.62 &0.8319 &0.8785 \\
			&IAA w/o ML & 3D Pose+Obj &87.03 &0.8458 & 0.8062\\
			&IAA w/o Atten & 3D Pose+Obj &89.38 &0.8904 &0.9168 \\
			&\textbf{IAA(ours)} & \textbf{3D Pose+Obj}& \textbf{93.65} & \textbf{0.9027} & \textbf{0.9341}\\
			
			\bottomrule
			\Xhline{0.5pt}
		\end{tabular}
	}
	\vspace{-1em}
\end{table}

We also compare the impact of various modules in the proposed Inversion-Interaction method on the success rate of \emph{Grasp} and \emph{Open drawer} tasks, which is shown in Table \ref{table-ablation2}. CIA-w/o-inversion represents that CIA randomly samples and generates trajectories from the latent space of the trained TrajGAN without initialization by GAN inversion. CIA-w/o-interactions only relies on the trajectory initialization without the subsequent policy improvement through interaction. CIA-w/o-IAAformer replaces the IAAformer with a video classifier to determine whether human videos and robot videos belong to the same task category as the reward. As is shown in Table \ref{table-ablation2}, CIA-w/o-inversion has difficulty in completing two tasks, since it's inefficient to optimize reasonable trajectories through a small amount of sampling in the vast latent space using two iterations. 
CIA-w/o-interactions performs better, which shows the necessity of using initialization to efficiently find the vicinity of the optimal trajectory.
CIA-w/o-IAAformer is also limited by the inappropriate objective function.
With all the components, CIA can achieve the success rate of 90\% in \emph{Grasp} and 87\% in \emph{Open drawer}.

\begin{table}[t]
	\setlength{\tabcolsep}{0.3em}
	
	\caption{Ablation experiments for each component of Inversion-Interaction method}
	\vspace{-0.1cm}
	\label{table-ablation2}
	\centering

	\footnotesize{
		\begin{tabular}{m{3.cm}<{\raggedright} m{2cm}<{\centering} m{2cm}<{\centering}}
		
			\Xhline{0.5pt}
			\toprule
			\textbf{Methods} &  \textbf{Grasp}  & \textbf{Open drawer}  \\
			\midrule
			CIA-w/o-inversion &0.27 & 0.17    \\
			CIA-w/o-interactions & 0.43 & 0.37  \\
			CIA-w/o-IAAformer & 0.37 &0.43\\
			
			\textbf{CIA} & \textbf{0.90}& \textbf{0.87}  \\
			
			\bottomrule
			\Xhline{0.5pt}
		\end{tabular}
	
	}
\end{table}

\subsection{Adaptation of CIA}

\begin{figure}[t]
	\centering
	\includegraphics[width=9cm]{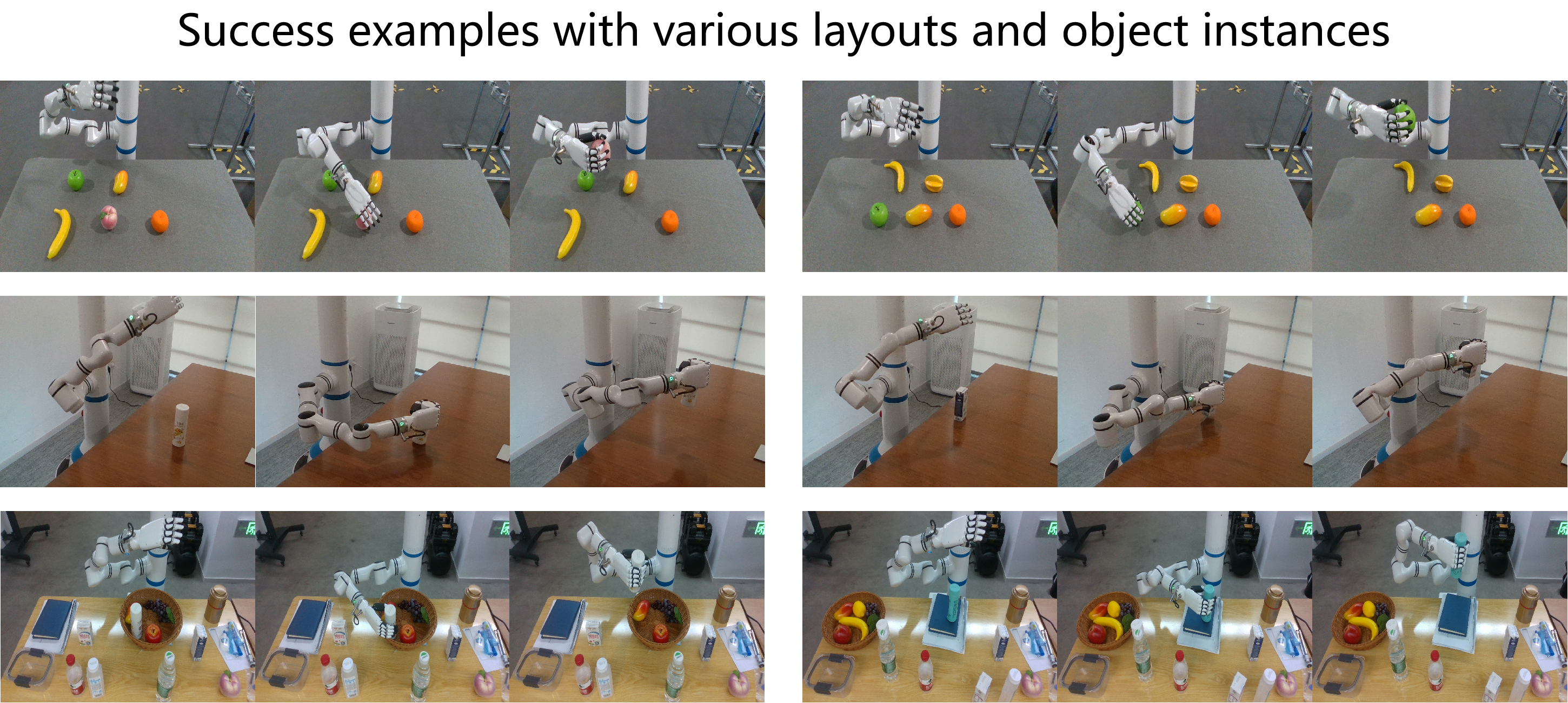}
	\caption{Some test trajectories with various object instances and scenario layouts.}
	\label{exp-various-fig}
	\vspace{-1em}
\end{figure}

\noindent\textbf{Adaptation to Novel Scenarios.} 
We have verified that our proposed CIA can achieve a success rate of 90\% in \emph{Grasp} task (Table \ref{table-cia}). Note that the object positions in our experiments are different from those in human videos. Some test trajectories with various scenario layouts are shown in Fig. \ref{exp-various-fig}. As is shown, CIA can adapt to various scenario layouts successfully. Specifically, CIA can grasp different object instances at different positions and heights. In addition, CIA takes advantage of the generalization of the visual foundational model \cite{liu2023grounding} to adapt to different backgrounds and contexts, as shown in the different rows of the figure.

\noindent\textbf{Adaptation to Novel Object Instances.} To verify the CIA's adaptability to diverse object instances, we randomly change the position of objects in experiments of different tasks. For the \emph{Grasp} task, CIA can successfully grasp various fruits, milk cartons, coffee bottles and other beverage bottles, which is shown in Fig. \ref{exp-task-pic-b} and \ref{exp-various-fig}. We used two different sizes of refrigerators and drawers for testing \emph{Open/Close refrigerator} and \emph{Open/Close drawer} tasks. In \emph{Wipe} task, we change the position of the words to be erased on the whiteboard. As is shown in Table \ref{table-cia}, CIA can adapt to different object instances with a high success rate.

\subsection{Failure Analysis}
\begin{figure}[t]
	\centering
	\includegraphics[width=6.8cm]{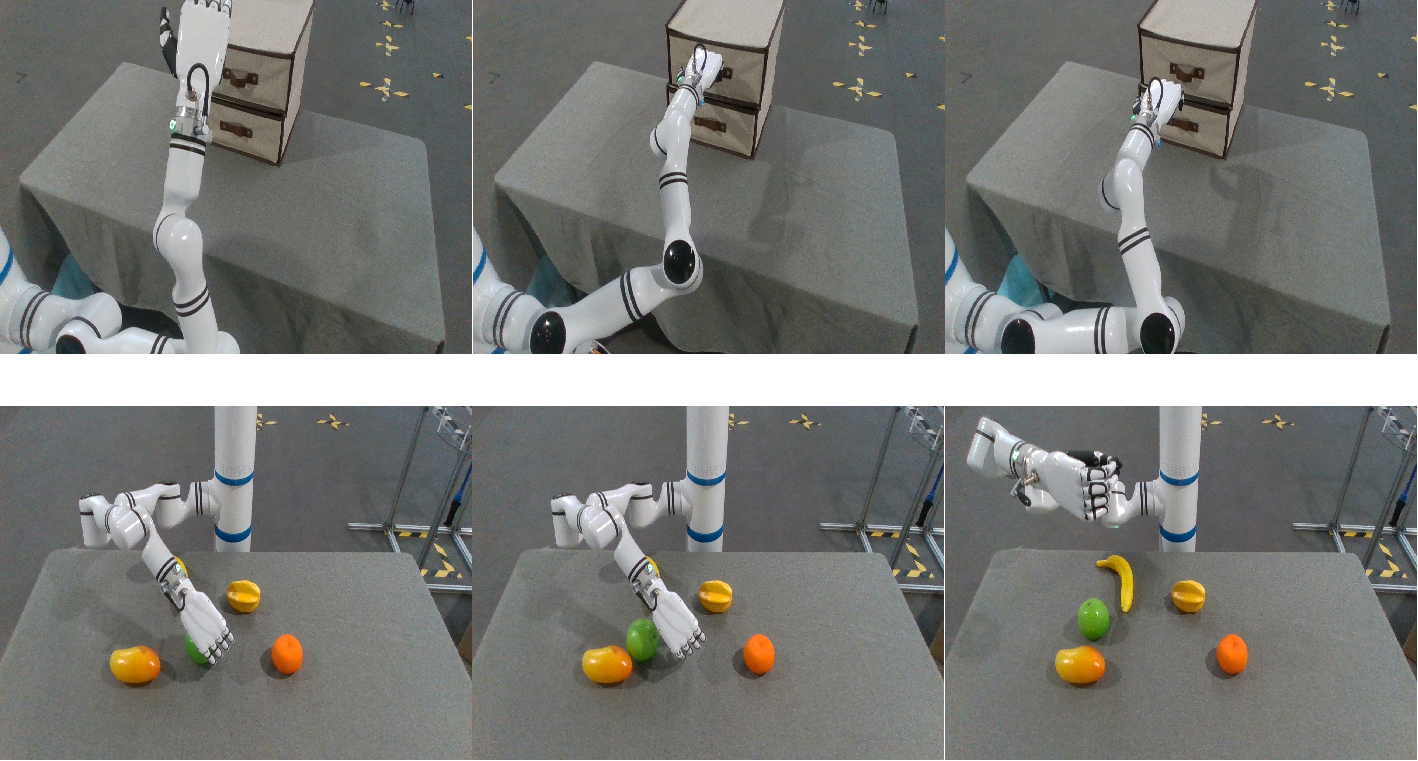}
	\caption{Some failure trajectories of \emph{Open drawer} and \emph{Grasp} tasks.}
	\label{exp-failure}
	\vspace{-1.em}
\end{figure}

Some failure examples of \emph{Open drawer} and \emph{Grasp} tasks are shown in Fig. \ref{exp-failure}. In the above example, the robot fails to grasp the drawer handle due to the inaccuracy between the contact points in the generated trajectory and environmental variables, i.e. the handle. 
In the example below, the apple slides out of the hand, which can be prevented by improving the accuracy of robot hand positions or postures.

\section{Conclusion}\label{section5}
We propose a three-stage real-world robot learning framework, denoted as Contrast-Imitate-Adapt (CIA), which can learn robotic skills from raw human videos. Leveraging advances in computer vision, task priors and action priors are extracted. 
Subsequently, an Interaction-Aware Alignment transformer (IAAformer) with attention mechanisms and a hindsight interaction contrastive loss is designed to learn task priors by temporally aligning two videos.
Moreover, Editable Movement Primitive (TrajGAN) is trained to generate trajectories to imitate action priors.
To adapt to novel scenarios different from the human videos, the Inversion-Interaction method is proposed to utilize GAN inversion in the latent space to initialize coarse trajectories and refine them through limited interaction. 
Notably, we introduce Cross Entropy Method based on Trajectory Semantic sampling (TS-CEM) for safe and efficient interaction, where the contrast results produced by IAAformer as rewards.
We empirically demonstrate that CIA can significantly outperform previous state-of-the-art methods in six everyday tasks.

Our work suggests several potential directions for future research.
In terms of learning task priors, while CIA utilizes a contrast between robot and human videos as an evaluation for robot execution, other methods, e.g. vision-language foundational models, can also be explored to achieve this functionality.
In terms of learning action priors, accomplishing more complex tasks, e.g. dealing with occluded objects, remains a challenge.
In addition, while CIA employs trajectory generation models for open-loop robot control, designing more robust models to generate real-time and reasonable trajectories for closed-loop control is another open issue.


\bibliographystyle{IEEEtran}
\bibliography{reference}




\par\noindent 
\parbox[t]{\linewidth}{
	\noindent\parpic{\includegraphics[height=0.88in,clip,keepaspectratio]{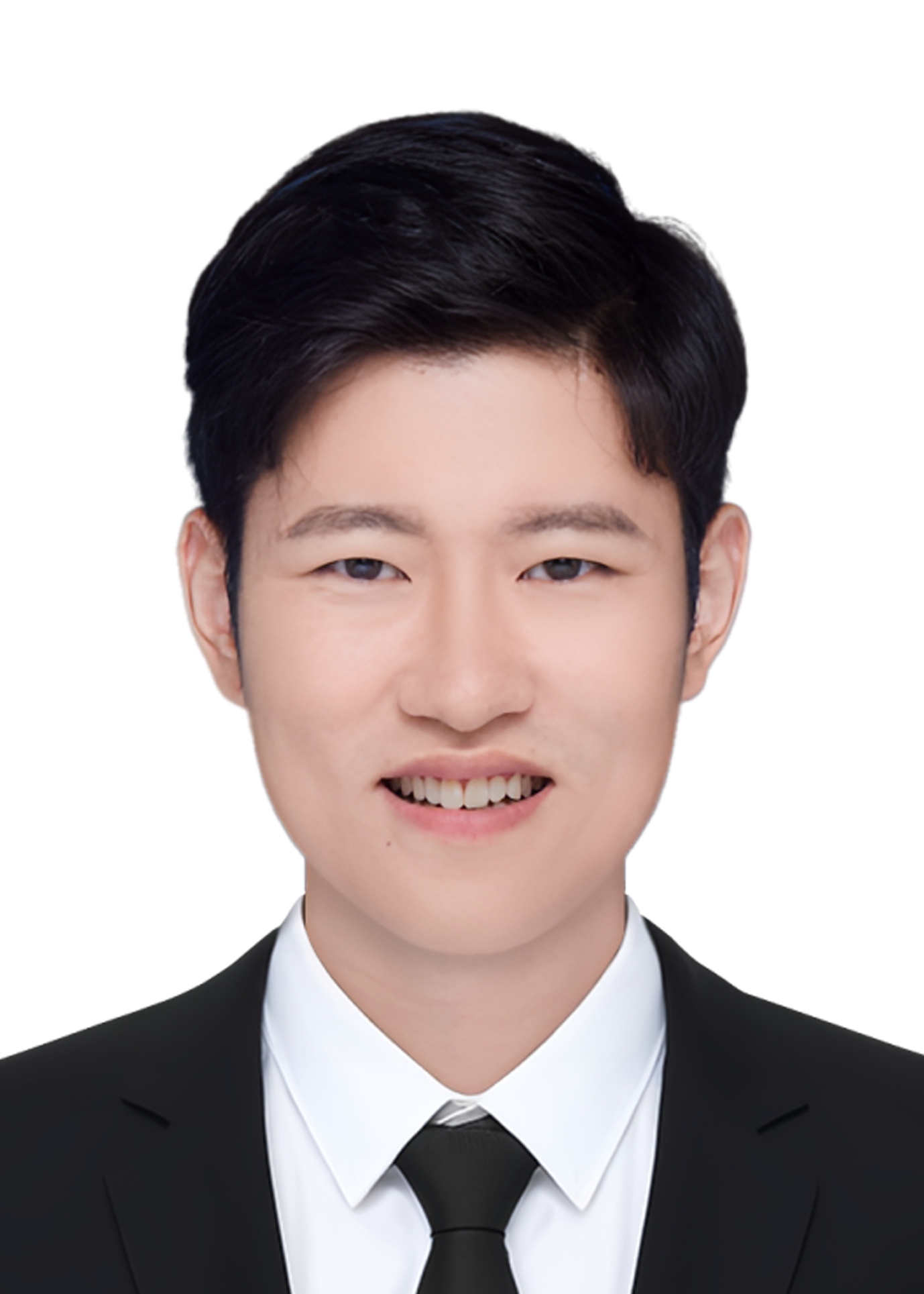}}
	\noindent {\bf Zhifeng Qian}\
	received the B.S. degree from Soochow University, China in 2018. He is currently working toward the Ph.D. degree in Tongji University, China. 
	His research interests include imitation learning, deep reinforcement learning.}
\vspace{0.5\baselineskip}

\par\noindent 
\parbox[t]{\linewidth}{
	\noindent\parpic{\includegraphics[height=0.88in,clip,keepaspectratio]{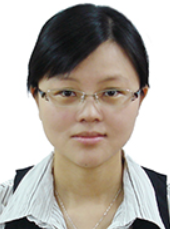}}
	\noindent {\bf Mingyu You}\
	(Member, IEEE) received the B.S. and Ph.D. degrees from Zhejiang University, China, in 2002 and 2007. She is currently a Professor in Tongji University. Her
	research interests include artificial intelligence and robotic learning.}
\vspace{0.5\baselineskip}


\par\noindent 
\parbox[t]{\linewidth}{
	\noindent\parpic{\includegraphics[height=0.88in,clip,keepaspectratio]{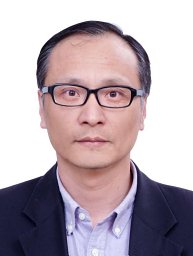}}
	\noindent {\bf Hongjun Zhou}\
	(Member, IEEE) received the B.S. degree from Dalian University of Technology, China, in 1994, and the M.S. and Ph.D. degrees from Chuo University, Japan, in 2001 and 2004. 
	He is currently an Associate Professor in Tongji University. His research interests include robot SLAM and robot imitation learning.}
\vspace{0.5\baselineskip}


\par\noindent 
\parbox[t]{\linewidth}{
	\noindent\parpic{\includegraphics[height=0.88in,clip,keepaspectratio]{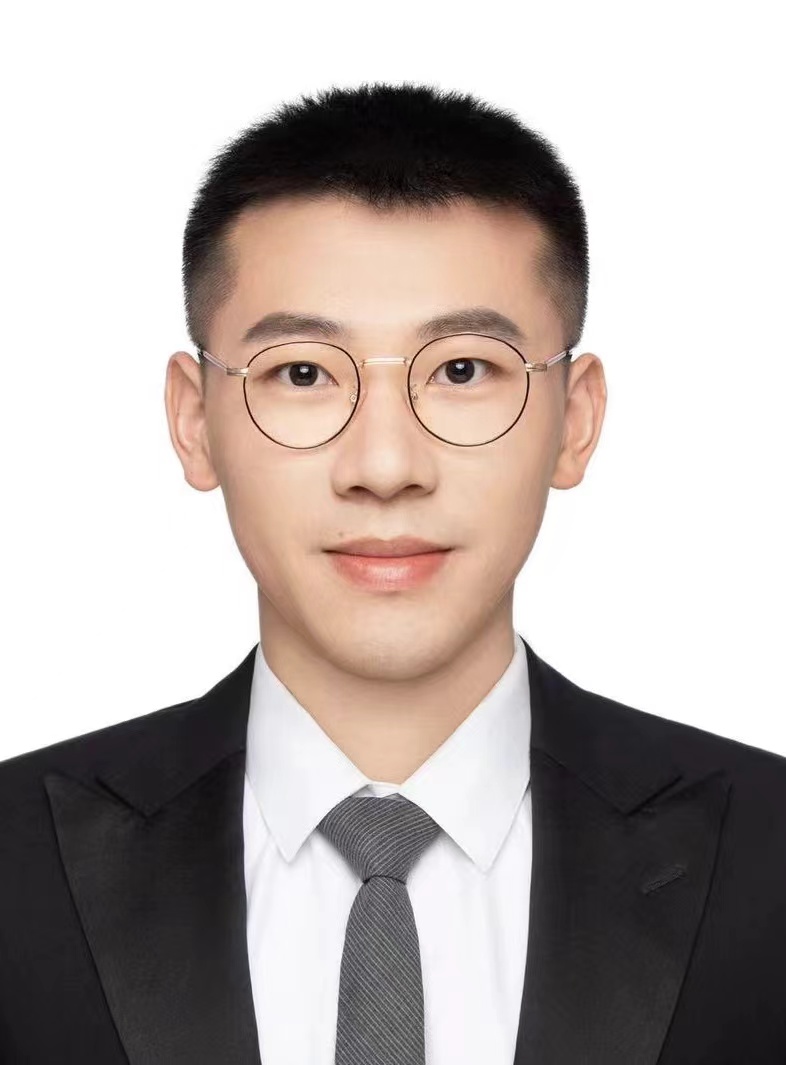}}
	\noindent {\bf Xuanhui Xu}\
	received the B.S. degree from Donghua University, China in 2018. He is currently working toward the Ph.D. degree in Tongji University. 
	His research interests include multimedia understanding and imitation learning.}
\vspace{0.5\baselineskip}

\par\noindent 
\parbox[t]{\linewidth}{
	\noindent\parpic{\includegraphics[height=0.88in,clip,keepaspectratio]{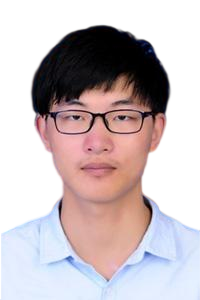}}
	\noindent {\bf Hao Fu}\
	 received the Master's degree from Tongji University, China in 2021. He is currently working toward the Ph.D. degree in Tongji University. His research interests include multi-agent system and deep reinforcement learning.}
\vspace{0.5\baselineskip}


\par\noindent 
\parbox[t]{\linewidth}{
	\noindent\parpic{\includegraphics[height=0.88in,clip,keepaspectratio]{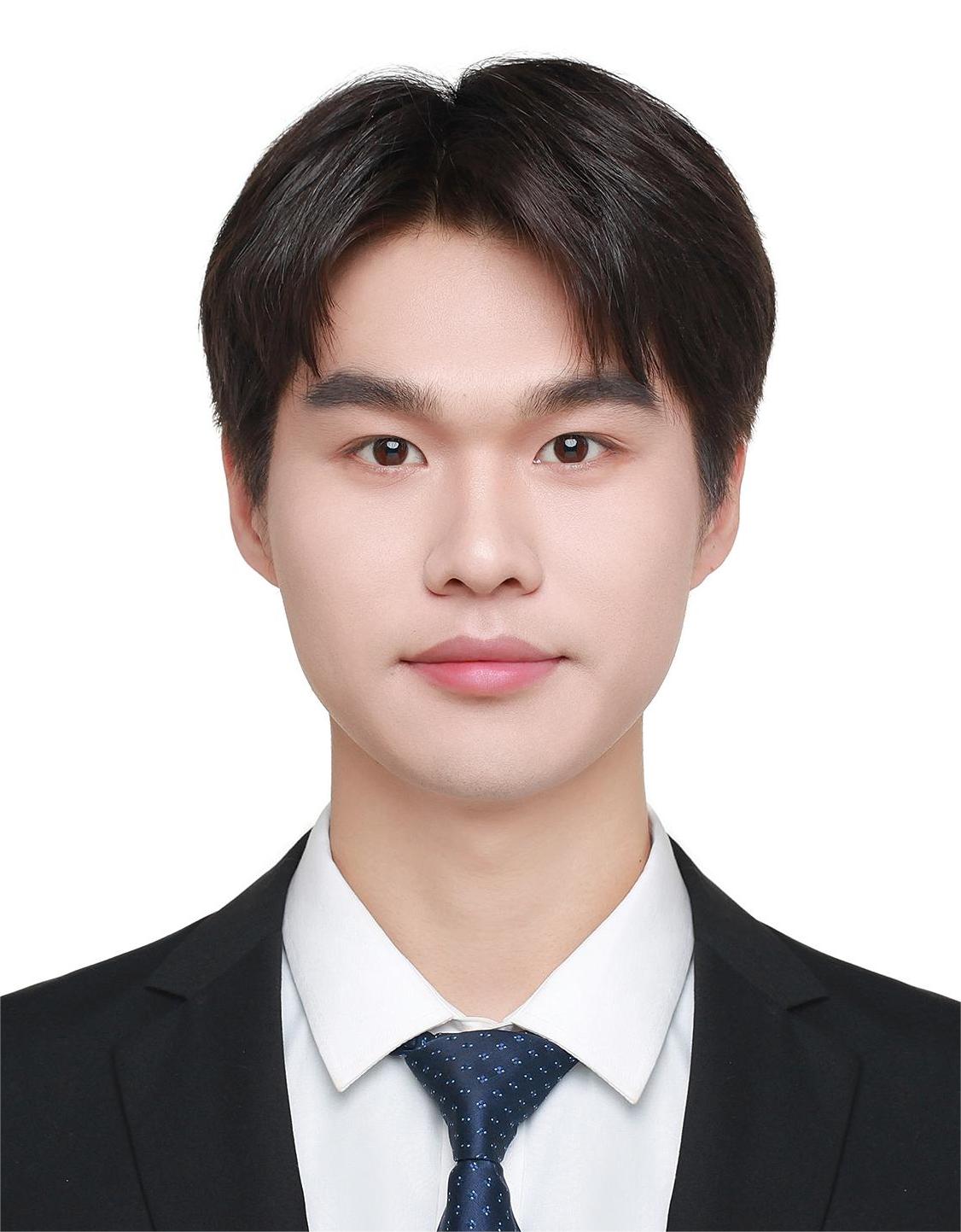}}
	\noindent {\bf Jinzhe Xue}\
	received the B.S. degree from Tongji University, China in 2022. He is currently working toward the M.S. degree with the College of Electronics and Information Engineering, Tongji University.}
\vspace{0.5\baselineskip}


\par\noindent 
\parbox[t]{\linewidth}{
	\noindent\parpic{\includegraphics[height=0.88in,clip,keepaspectratio]{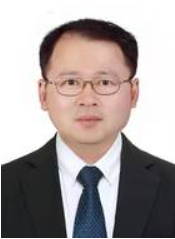}}
	\noindent {\bf Bin He}\
	(Member, IEEE) received the B.S. degree from Jilin University, Changchun, China, in 1996, and the Ph.D. degree from Zhejiang University, China, in 2001.
	He is currently a Professor in Tongji University, Shanghai, China. His current research interests include intelligent robot control and biomimetic microrobots.}


\vfill

\end{document}